\documentclass[10pt,journal,cspaper,compsoc]{IEEEtran}
\ifCLASSOPTIONcompsoc
\else
  \usepackage{cite}
\fi
\ifCLASSINFOpdf
\else
\fi

\usepackage{graphicx}
\usepackage{amsmath}
\usepackage{amsthm}
\usepackage{color}
\usepackage{times}
\usepackage{epsfig}
\usepackage{amssymb}
\usepackage{wasysym}
\usepackage{verbatim}
\usepackage{multirow}
\usepackage{makecell}
\usepackage{caption}
\usepackage{subcaption}
\usepackage{array}
\usepackage[]{algorithm2e}
\usepackage{algpseudocode}
\newcolumntype{L}[1]{>{\raggedright\let\newline\\\arraybackslash\hspace{0pt}}m{#1}}
\newcolumntype{C}[1]{>{\centering\let\newline\\\arraybackslash\hspace{0pt}}m{#1}}
\newcolumntype{R}[1]{>{\raggedleft\let\newline\\\arraybackslash\hspace{0pt}}m{#1}}

\def\mQ{\mathcal{Q}}

\def\1n{\mathbf{1}_n}
\def\0{\mathbf{0}}
\def\1{\mathbf{1}}

\def\A{{\bf A}}

\def\D{{\bf D}}

\def\F{{\bf F}}

\def\I{{\bf I}}
\def\J{{\bf J}}
\def\K{{\bf K}}
\def\L{{\bf L}}

\def\S{{\bf S}}

\def\U{{\bf U}}

\def\W{{\bf W}}

\def\Y{{\bf Y}}

\def\f{{\bf f}}

\def\p{{\bf p}}

\def\x{{\bf x}}

\begin{document}
%
\title{Long-Term Identity-Aware Multi-Person Tracking for Surveillance Video Summarization}
%
%
%
%

\author{
Shoou-I Yu, Yi Yang, Xuanchong Li, and Alexander G. Hauptmann
}

%
%

\markboth{ }%
{}
%


\IEEEcompsoctitleabstractindextext{%
\begin{abstract}
	Multi-person tracking plays a critical role in the analysis of surveillance video. However, most existing work focus on shorter-term (e.g. minute-long or hour-long) video sequences. Therefore, we propose a multi-person tracking algorithm for very long-term (e.g. month-long) multi-camera surveillance scenarios. Long-term tracking is challenging because 1) the apparel/appearance of the same person will vary greatly over multiple days and 2) a person will leave and re-enter the scene numerous times. To tackle these challenges, we leverage face recognition information, which is robust to apparel change, to automatically reinitialize our tracker over multiple days of recordings. Unfortunately, recognized faces are unavailable oftentimes. Therefore, our tracker propagates identity information to frames without recognized faces by uncovering the appearance and spatial manifold formed by person detections. We tested our algorithm on a 23-day 15-camera data set (4,935 hours total), and we were able to localize a person 53.2\% of the time with 69.8\% precision. We further performed video summarization experiments based on our tracking output. Results on 116.25 hours of video showed that we were able to generate a reasonable visual diary  (i.e. a summary of what a person did) for different people, thus potentially opening the door to automatic summarization of the vast amount of surveillance video generated every day.

\end{abstract}

}

\maketitle

\IEEEdisplaynotcompsoctitleabstractindextext

%
\IEEEpeerreviewmaketitle

\section{Introduction}

\IEEEPARstart{S}{urveillance} cameras have been widely deployed to enhance safety in our everyday lives.
The recorded footage can further be used to analyze long term trends in the environment.
Unfortunately, manual analysis of large amounts of surveillance video is very difficult, thus motivating
the development of computational analysis of surveillance video.
A common first step of computational analysis is to track each person in the scene, which 
has led to the development of many multi-object tracking algorithms \cite{zhang2008global, collins2012multitarget, dctracker}.
However, two important points are largely neglected in the literature: 1) the usage of
identity information such as face recognition or any other cue that can identify an individual,
and 2) the exploration of real-world applications based on tracking output from hundreds or thousands of hours of surveillance video.

There are two main advantages of utilizing face recognition information for tracking.
First, face recognition empowers the 
tracker to relate a tracked person to a real-world living individual,
thus enabling individual-specific activity analysis.
Second, face recognition is robust to appearance/apparel change, thus making it well-suited
for tracker reinitialization in very long-term (e.g. month-long) surveillance scenarios.

We propose an identity-aware tracking algorithm
as follows.
Under the tracking-by-detection framework \cite{Okuma04aboosted}, the tracking task can be viewed as assigning each person detection
to a specific individual/label.
Face recognition output can be viewed as label information.
However, as face recognition is only available in a few frames, we propagate face recognition labels to other frames
using a manifold learning approach, which captures the appearance similarities and spatial-temporal layout of person detections.
The manifold learning approach is formulated as a constrained quadratic optimization problem and solved with nonnegative matrix optimization techniques.
The constraints included are the
mutual exclusion and spatial locality constraints which constrain the final solution to deliver a reasonable multi-person tracking output.

\begin{figure}[tp]
\centering
\includegraphics[scale=0.22]{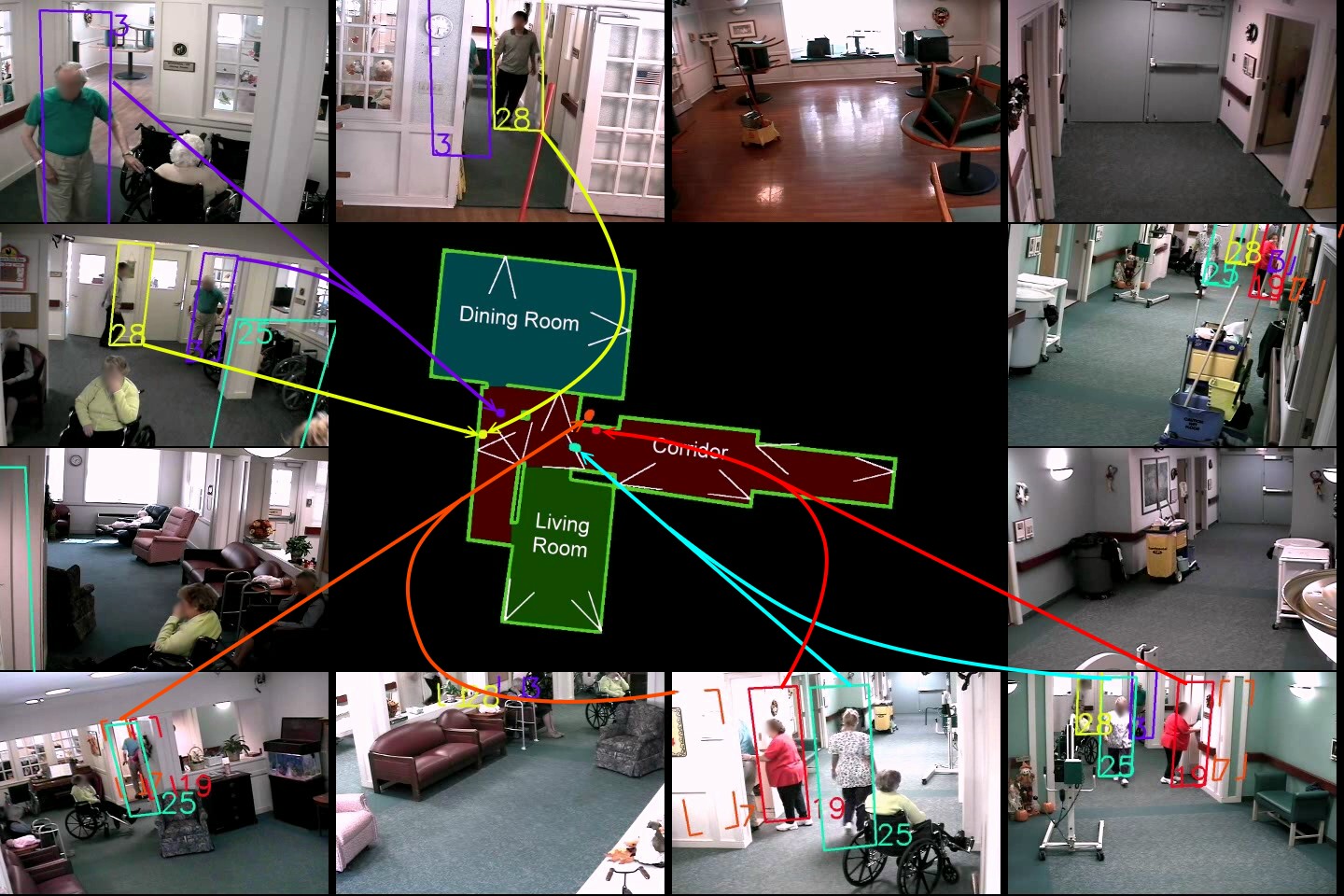}
\caption{
	The Marauder's Map for a nursing home (\emph{Caremedia Short} sequence \cite{Marauders}) with the map in the middle.
Dots on the map show the locations of different people.
The surrounding images are the views from each surveillance
camera. White lines correspond to the field-of-view of each camera. 
}
\label{fig:marauders}
\end{figure}

We performed tracking experiments on challenging data sets, including a 4,935 hour complex indoor tracking data set.
Our long-term tracking experiments showed that 
our method was effective in localizing and tracking each individual in thousands of hours of surveillance video.
An example output of our algorithm is shown in Figure~\ref{fig:marauders}, 
which shows the location of each identitifed person on the map in the middle of the image.
This is analogous to the \emph{Marauder's Map} described in the Harry Potter book series \cite{HarryPotter}.

\begin{figure}[tp]
\centering
\includegraphics[scale=0.28]{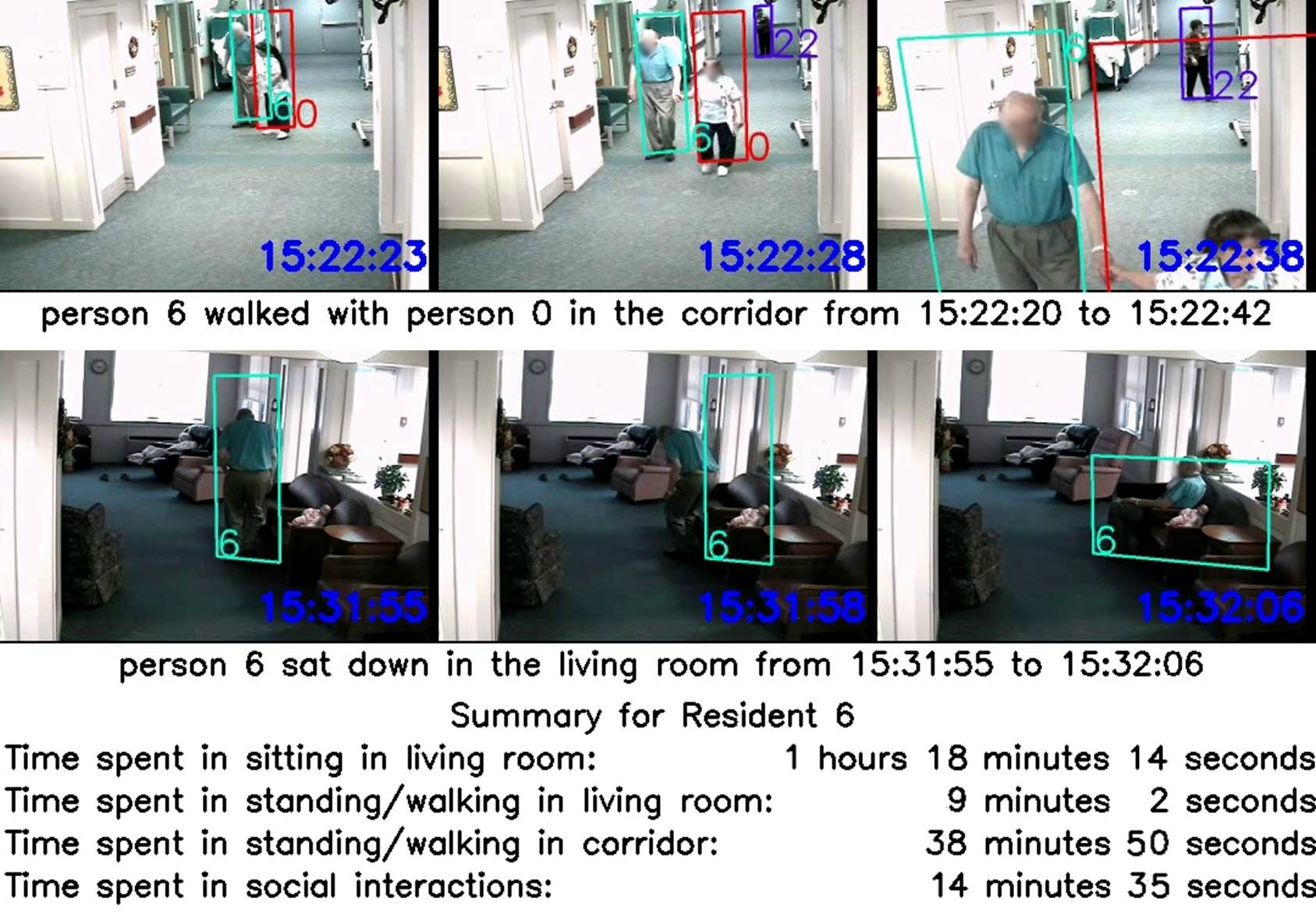}
\caption{
An example visual diary for an elderly resident in a nursing home.
The automatically generated textual description and snapshots are shown for the two events.
Long-term statistics are also shown.
}
\label{fig:diary11}
\end{figure}

To explore the utility of long-term multi-person tracking,
we performed \emph{summarization-by-tracking} experiments 
to acquire the \emph{visual diary} of a person. 
Visual diaries provide a person-specific summary of surveillance video
by showing snapshots and textual descriptions of the activities performed by the person.
An example visual diary of a nursing home resident is shown in Figure~\ref{fig:diary11}.
Experiments conducted on 116.25 hours of video show that we were able to summarize surveillance video with reasonable accuracy,
which further demonstrates the effectiveness of our tracker.

In sum, the main contributions of this paper are as follows:
\begin{enumerate}
\item 
We propose an identity-aware multi-object tracking algorithm. 
Our tracking algorithm leverages identity information which is utilized as sparse label information in a manifold learning framework.
The algorithm is formulated as a constrained quadratic optimization problem and solved with
nonnegative matrix optimization.
\item
A 15-camera multi-object tracking data set consisting of 4,935 hours of nursing home surveillance video
was annotated. This real-world data set enables us to perform very long-term tracking experiments to better
assess the performance and applicability of multi-object trackers.
\item
Video summarization experiments based on tracking output were performed on 116.25 hours of video. 
We demonstrate that the visual diaries generated from tracking-based summarization
can effectively summarize hundreds of hours of surveillance video.
\end{enumerate}

\section{Related Work}
\label{sec:related}




As multi-object tracking is a very diverse field, we only review work
that follows the very popular tracking-by-detection paradigm \cite{Okuma04aboosted}, which is also used
in our work.
For a more comprehensive and detailed survey we refer the readers to \cite{MOTSurvey}.

The tracking-by-detection paradigm has four main components:
object localization, appearance modeling, motion modeling and data association.
The object localization component generates a set of object location hypotheses for each frame. 
The localization hypotheses are usually noisy and contain false alarms
and misdetections, so
the task of the data association component
is to robustly group the location hypotheses which belong to the same physical object to form many different object trajectories.
The suitability of the grouping can be scored according to the coherence of the object's appearance and the smoothness of the object's motion,
which correspond to appearance modeling and motion modeling respectively.
We now describe the four components in more detail.

\subsection{Object Localization}
There are mainly three methods to find location hypotheses: using background subtraction,
using object detectors, and connecting single-frame detection results into tracklets.
The Probabilistic Occupancy Map (POM, \cite{POM}) combines background subtraction information from multiple cameras to jointly 
locate multiple objects in a single frame. 
Utilizing object detector output is one of the most common ways to localize tracking targets 
\cite{Okuma04aboosted, zamir2012gmcp, Marauders, pirsiavash2011globally, zhang2008global, andriyenko2011multi, dctracker,dettrajexclusion, collins2012multitarget,butt2013multi}.
The object detector is run on each frame of the video, and the detection results
serve as the location hypotheses for subsequent processing.
Localized objects in each frame could be connected to create \emph{tracklets} 
\cite{hiertracklets, oldams, kuoidentity, ramCRF, hybridboost, yang2012multi, TMCNF}, which are
short tracks belonging to the same physical object. 
Tracklets are usually formed in a very conservative way to avoid connecting two physically different objects.

\subsection{Appearance Models}
Appearance models discriminate between detections belonging to the same physical object and other objects.
Color histograms \cite{ILP, leibe2007coupled, zhang2008global, hiertracklets, hybridboost, hexagonal, zamir2012gmcp, Marauders, TMCNF} have been widely used to represent the appearance of objects,
and the similarity of the histograms is often computed with the Bhattacharyya distance \cite{zhang2008global, hexagonal}.
Other features such as Histogram of Oriented Gradients \cite{HOG} have also been used \cite{oldams, kuoidentity}.

Appearance models can also be learned from tracklets. 
The main assumption of tracklets is that all detections in a tracklet belong to the same object,
and \cite{oldams, ramCRF, yang2012multi, bae14robust, wang2014tracklet, kuoidentity} exploit this assumption to learn more discriminative appearance models.
Note that the ``identity'' in our work is different from \cite{kuoidentity}, which utilized person re-identification techniques to improve the appearance model.
We, however, focus on the ``real-world identity'' of the person, which is acquired from face-recognition.

Appearance models based on incremental manifold/subspace learning 
has also been utilized in previous work \cite{zhang2007graph,hu2012single,salti2012adaptive} to learn subspaces for appearance features
that can better differentiate tracked targets and background in single or multi-object tracking.
However, \cite{hu2012single} utilized multiple \emph{independent} particle filters, 
which may have the issue of one particle filter ``hijacking'' the tracking target of another particle filter \cite{khan2005mcmc,hess2009discriminatively}.
Our method alleviates this issue as 
we \emph{jointly} optimize for all trajectories to acquire a more reasonable set of trajectories.

\subsection{Motion Models}
Objects move in a smooth manner, and motion models can capture this assumption to better
track objects.
\cite{zhang2008global, KSP, pirsiavash2011globally, Marauders, TMCNF} use the bounded velocity model to model motion, i.e. an object cannot move faster than a given velocity.
\cite{leibe2007coupled, zamir2012gmcp, butt2013multi} improve upon this by 
modeling motion with the constant velocity model, which is able to model acceleration.
Higher order methods such as spline-based methods \cite{collins2012multitarget, dctracker} and the Hankel matrix \cite{DicleICCV13} 
can model even more sophisticated motions. 
\cite{yang2012multi} assumes that different objects in the same scene move in similar but potentially non-linear ways, 
and the motion of highly confident tracklets can be used to infer the motion of non-confident tracklets.

\subsection{Data Association}
A data association algorithm takes the object location hypotheses, appearance model and motion model as input
and finds a disjoint grouping of the object location hypotheses which best describes the motion of objects in the scene.
Intuitively, the algorithm will decide whether to place two object location hypotheses in the same group based on their \emph{affinity},
which is computed from the appearance and motion models.

The Hungarian algorithm and the network flow are two popular formulations.
Given the pair-wise affinities, the Hungarian algorithm can find 
the optimal matching between two sets of object location hypotheses  
in polynomial time \cite{hiertracklets,hybridboost,kuoidentity,oldams,collins2012multitarget}.
In the network flow formulation \cite{zhang2008global, pirsiavash2011globally, KSP, wang2014tracking}, 
each path from source to sink corresponds to the trajectory of an object.

Many trackers have been formulated as a general Integer Linear Programming (ILP) 
problem.
\cite{ILP,TMCNF,hexagonal} solved the ILP by first relaxing the integral constraints to continuous constraints
and then optimizing a Linear Program.
\cite{ferrari2001real,marin2014detecting} formulated tracking as clique partitioning,
which can also be formulated as an ILP problem and solved by a heuristic clique merging method.

More complex data association methods have also been used, including
continuous energy minimization \cite{andriyenko2011multi}, 
discrete-continuous optimization \cite{dctracker},
Block-ICM \cite{collins2012multitarget}, 
conditional random fields \cite{ramCRF,dettrajexclusion},
generalized minimum clique \cite{zamir2012gmcp}
and quadratic programming \cite{leibe2007coupled,chari2015pairwise}.

However, it is non-trivial to incorporate identity information such as face recognition 
into the aforementioned methods.
One quick fix may be to assign identities to trajectories
after the trajectories have been computed. 
However, problems occur if there are identity-switches in a single trajectory.
Another method proposed by \cite{POM} utilized the Viterbi algorithm to find a trajectory which passes through all the identity observations
of each person.
However, Viterbi search cannot be performed simultaneously over all individuals, 
and \cite{POM} had to performed Viterbi search sequentially, i.e. one individual after another.
This greedy approach lead to ``hijacking'' of another person's trajectory \cite{POM}, which is not ideal.
Therefore, to achieve effective identity-aware tracking, it is ideal to
design a data association framework which can directly incorporate identity information into the optimization process.

\subsubsection*{Identity-Aware Data Association}

Previously proposed data association methods \cite{TMCNF, zervos2013multi}, \cite{lu2013learning}, \cite{dehghan2015target} and \cite{Marauders} utilized identity information for tracking.
There have been other work which utilized transcripts from TV shows to perform face recognition and identity-aware face tracking \cite{everingham2009taking,sivic2009you}, but this is not the main focus of our paper.

\cite{TMCNF, zervos2013multi} formulated identity-aware tracking as an ILP 
and utilized person identification information from numbers written on an athlete's jersey or from face recognition.
\cite{TMCNF,zervos2013multi} utilized a global appearance term as their appearance model to assign identities to detections.
However, the global term assumes a fixed appearance template for an object,
which may not be applicable in long surveillance recordings as the appearance of the same person may change.

\cite{lu2013learning} utilized a few manually labeled training examples and play-by-play text in a Conditional Random Field formulation to accurately track and identify sports players.
However, this method may not work as well in surveillance domains where play-by-play text is not available.

\cite{dehghan2015target} utilized online structured learning to learn a target-specific appearance model, 
which is used 
in a network flow framework. However, \cite{dehghan2015target} 
utilized densely-sampled windows instead of person bounding boxes as input, which may be too time-consuming to compute
in long videos.


\cite{Marauders} utilized face-recognition as sparse label information in a semi-supervised tracking framework.
However, \cite{Marauders} 
does not incorporate the spatial locality constraint into the optimization step,
which might lead to solutions showing a person being at multiple
places at the same time. This becomes very severe in crowded scenes.
Also, the method needs a Viterbi search to compute the final trajectories.
The Viterbi search requires the start and end locations of all trajectories, which is an
unrealistically restrictive assumption for long-term tracking scenarios.
In this paper, we enhance this tracker by adding the spatial-locality constraint term, which enables tracking in crowded scenes
and also removes the need for the start and end locations of a trajectory.

\section{Methodology}
\label{sec:tracking}

\begin{figure*}[tp]
\centering
\begin{subfigure}{0.48\textwidth}
\centering
\includegraphics[scale=0.40]{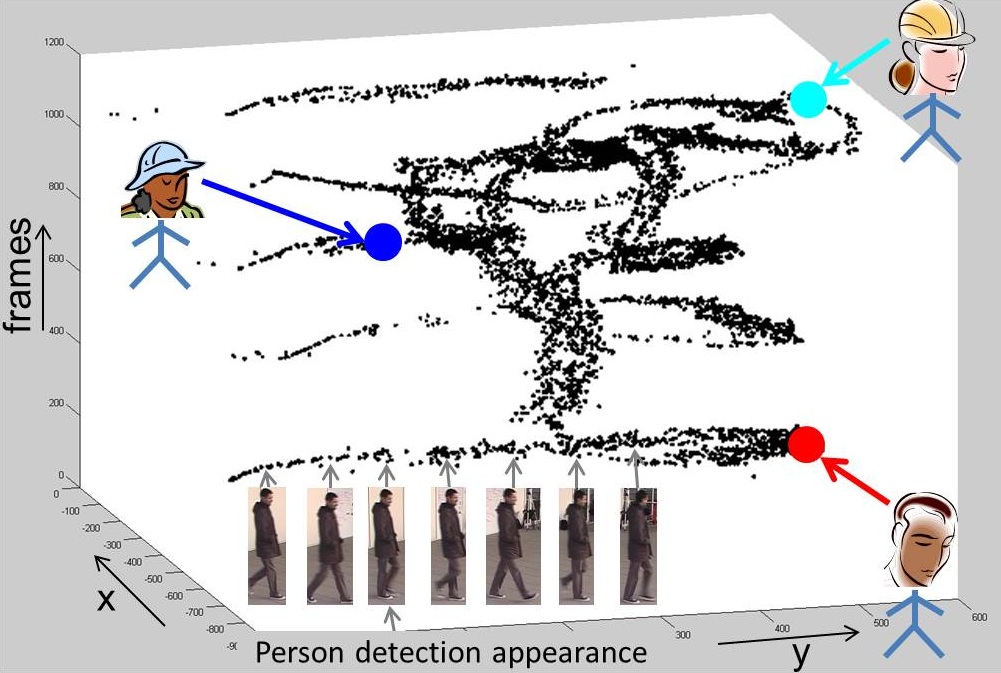}
\caption{Input to tracking algorithm: location and appearance of person detection plus recognized faces for some person detections.}\label{fig:man1}
\end{subfigure}
\begin{subfigure}{0.48\textwidth}
\centering
\includegraphics[scale=0.27]{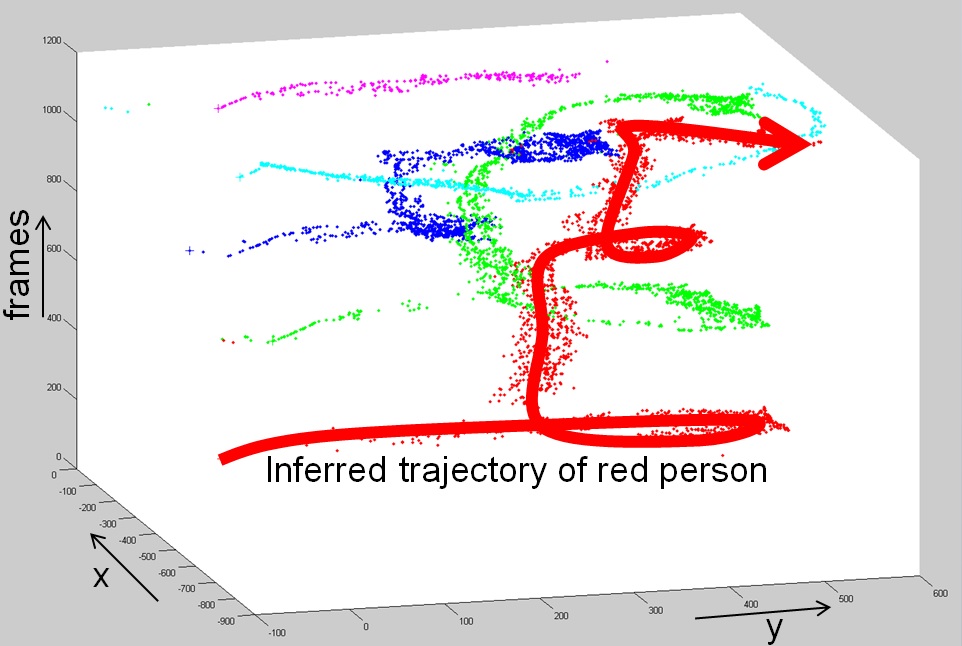}
\caption{Output of tracking algorithm: partitioning of the person detections into different trajectories.}\label{fig:man2}
\end{subfigure}
\caption{
Illustration of the input and output of our tracking algorithm. 
Each person detection is a point in the $(x, y, t)$ space. 
The $z$ axis is irrelevant in this case, because
the figures are drawn based on the person detections from the \emph{terrace1} data set \cite{POM},
where people walk on a common ground plane.
Best viewed in color.
}
\label{fig:manifold}
\end{figure*}

Tracking-by-detection-based multi-object tracking can be viewed as a constrained clustering problem as 
shown in Figure~\ref{fig:manifold}.
Each location hypothesis, which is a person detection result, can be viewed as a point in the spatial-temporal space, and our goal
is to group the points so that the points in the same cluster belong to a single trajectory.
A trajectory should follow the mutual exclusion constraint and spatial-locality constraint, which are defined as follows.
\begin{itemize}
\item \textbf{Mutual Exclusion Constraint}: a person detection result can only belong to at most one trajectory. 
\item \textbf{Spatial-Locality Constraint}: two person detection results belonging to a single trajectory should be reachable
with reasonable velocity, i.e. a person cannot be in two places at the same time. 
\end{itemize}
Sparse label information acquired from sources such as face recognition can be used to assign real-world 
identities and also enhance tracking performance. 

Our tracking algorithm has three main steps.
\begin{enumerate}
\item Manifold construction based on appearance and spatial affinity: 
The appearance and spatial affinity
respectively assumes that 1) similar looking person detections are likely to be of the same individual and 2) person detections
which are spatially and temporally very close to each other are also likely to be of the same individual.
\item Spatial locality constraint: This constraint encodes the fact that a person cannot be at multiple places at the same time.
	In contrast to the manifold created in the previous step which encodes the \emph{affinity} of two person detections,
	this constraint encodes the \emph{repulsion} of two person detections.
\item Constrained nonnegative optimization: Our nonnegative optimization method
acquires a solution which simultaneously 
satisfies the manifold assumption, the mutual exclusion constraint and the spatial-locality constraint.
\end{enumerate}
In the following sections, we first define our notations, then the 3 aforementioned steps are detailed.

\subsection{Notations}
In this paper, given a matrix $\A$, let $\A_{ij}$ denote the element on the $i$-th row and $j$-th column of $\A$.
Let $\A_i$ denote the $i$-th row of $\A$.
$Tr(\cdot)$ denotes the trace operator.
$\left|\cdot \right|_F$ is the Frobenius norm of a matrix.
Given an positive integer $m$, $\textbf{1}_{m} \in \mathbb{R}^{m}$ is a column vector with all ones.

Hereafter, we call a person detection result an \emph{observation}.
Suppose the person detector detects $n$ observations.
Let $c$ be the number of tracked individuals,
which can be determined by either a pre-defined gallery of faces or
the number of unique individuals identified by the face recognition algorithm.
Our task is to assign a class label to each observation.
Let $\F \in \{0, 1\}^{n \times c}$ be the
label assignment matrix for all observations.
Without loss of generality, $\F$
is reorganized such that the observations from
the same class are located in consecutive rows, i.e.
the $j$-th column of $\F$ is given by:
\begin{equation}
\label{eq:Fdef}
\F_{*j} = [\underbrace{0,\dots,0}_{\sum_{i=1}^{j-1}m^{(i)}},
	\underbrace{1,\dots,1}_{m^{(j)}},
	\underbrace{0,\dots,0}_{\sum_{i=j+1}^{c}m^{(i)}}]^T,
\end{equation}
where $m^{(j)}$ is the number of observations in the $j$-th class.
If the $p$-th element in $\F_{*j}$, i.e. $\F_{pj}$, is 1, 
it indicates that the $p$-th observation 
corresponds to the $j$-th person.
According to Equation~\ref{eq:Fdef}, it can be verified that 
\begin{equation}
\label{eq:FTFI}
\F^T\F=\begin{bmatrix}\F_{*1}^T \\ \vdots \\ \F_{*c}^T\end{bmatrix}\begin{bmatrix}\F_{*1} & \dots & \F_{*c}\end{bmatrix}=\textrm{diag}\left(\begin{bmatrix} m^{(1)} \\ \vdots \\ m^{(c)} \end{bmatrix}\right)=\J.
\end{equation} 
The $i$-th observation is described by 
a $d$ dimensional color histogram $\x^{(i)} \in \mathbb{R}^d$,
frame number $t^{(i)}$, and 3D location $\p^{(i)} \in \mathbb{R}^3$ which corresponds to the 3D location of the bottom center of the bounding box.
In most cases, people walk on the ground plane, and the $z$ component becomes irrelevant.
However, our method is not constrained to only tracking people on the ground plane.

\subsection{Manifold Construction based on Appearance and Spatial Affinity}
\label{sec:laplacian}

There are two aspects we would like to capture with manifold learning: 1) appearance affinity and 2) spatial affinity,
which we will detail in the following sections.

\subsubsection{Modeling Appearance Affinity}

\begin{figure*}[tp]
\centering
\includegraphics[scale=0.29]{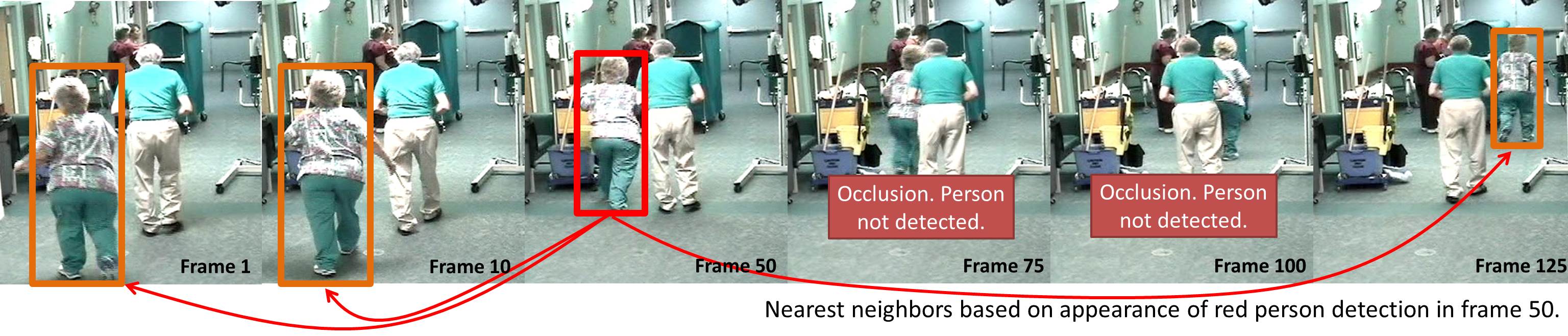}
\caption{
Intuition of appearance-based nearest neighbor selection.
The nearest neighbors for the red person detection in frame 50 are shown.
No nearest neighbors are found in frames 75 and 100 as the person is occluded.
Nevertheless, once the person is no longer occluded, the nearest neighbor connections could be made again,
thus overcoming this occlusion.
}
\label{fig:neighbor}
\end{figure*}

\label{subsec:neighbor}

Based on the assumption that two observations with similar appearance are likely to belong to the same individual,
we build the manifold structure by finding
nearest neighbors for each observation.
Observation $j$ is qualified to be a nearest neighbor of observation $i$ if 1) $j$ is
reachable with reasonable velocity, i.e. $v^{(ij)} \le V$, 
2) $i$ and $j$ should not be too far apart in time, i.e. $ \left| t^{(i)} - t^{(j)} \right| \le T$, and
3) both observations should look similar, i.e. the similarity of color histograms $\x^{(i)}$ and $\x^{(j)}$ should be larger than a threshold $\tau$.
We define 
$
v^{(ij)}=\frac{ \max \left( ||\p^{(i)}-\p^{(j)}||_2 - \delta, 0 \right) }{ |t^{(i)} - t^{(j)}| + \epsilon }.$
$\epsilon$ is a small number to avoid division by zero.
$\delta$ models the maximum localization error of the same person between different cameras due to calibration and person detection errors.
$V$ is the maximum velocity a person can achieve.
$T$ limits how far we look for nearest neighbors in the time axis.
The similarity between two histograms is computed with the exponential-$\chi^2$ metric:
$\chi^2( \x^{(i)}, \x^{(j)} ) = exp\left( -\frac{1}{2} \sum_{l=1}^d \frac{ \left( \x^{(i)}_l - \x^{(j)}_l \right)^2 }{ \x^{(i)}_l + \x^{(j)}_l } \right).$
For observation $i$, let $\mQ^{(i)}$ be the set of up to $k$ most similar observations which satisfy the three aforementioned criteria.
We can then compute the sparse affinity matrix $\W \in \mathbb{R}^{n \times n}$ as follows.
If $j \in \mQ^{(i)}$, then $\W_{ij} = \chi^2 \left( \x^{(i)}, \x^{(j)} \right)$.
Otherwise $\W_{ij}=0$.
The diagonal degree matrix $\D$ of $\W$ is computed, i.e. $\D_{ii} = \sum_{l=1}^{n} \W_{il}$.
Then, the Laplacian matrix which captures the manifold structure in the appearance space is $\L = \D - \W$.

This method of finding neighbors makes our tracker more robust to occlusions.
Occlusions may cause the tracking target to be partially or completely
occluded.  However, the tracking target usually reappears after a few frames.
Therefore, instead of trying to explicitly model occlusions, we try to connect
the observations of the tracking target before and after the occlusion.
As demonstrated in Figure~\ref{fig:neighbor}, despite heavy occlusions in a time segment, the algorithm
can still link the correct detections after the occlusion.
The window size $T$ affects the tracker's ability to recover from occlusions.
If $T$ is too small, the method will have difficulty recovering from occlusions that last longer than $T$.
However, a large $T$ may increase chances of linking two different objects.

\subsubsection{Modeling Spatial Affinity}

Other than modeling person detections of similar appearance,
person detections which are a few centimeters apart in the same or neighboring frames are also very likely to belong to the same person.
This assumption is reasonable in a multi-camera scenario because
multiple detections will correspond to the same person, and due to calibration and person detection errors, not all detections will be projected
to the exact same 3D location.
Therefore, regardless of the appearance difference which may be resulting from non-color-calibrated cameras,
these detections should belong to the same person.
We therefore encode this information with another Laplacian matrix $\K \in \mathbb{R}^{n \times n}$
defined as follows.
Let $\mathcal{K}^{(i)}$ be the set of observations which are less than distance $\tilde{\delta}$ away and less than $\tilde{T}$ frames away from observation $i$.
We compute the affinity matrix $\A \in \mathbb{R}^{n \times n}$ from $\mathcal{K}^{(i)}$ 
by setting $\A_{ij} = 1$ if $j \in \mathcal{K}^{(i)}$ and $\A_{ij} = 0$ otherwise.
Define $\hat{\D} \in \mathbb{R}^{n \times n}$ as a diagonal matrix where $\hat{\D}_{ii}$ is the sum of $\A$'s $i$-th row.
Following \cite{ng2002spectral}, 
the normalized Laplacian matrix is computed: $\K = \I - \hat{\D}^{-\frac{1}{2}}\A\hat{\D}^{-\frac{1}{2}}$.
The parameters $\tilde{\delta}$ and $\tilde{T}$ for spatial affinity should be set more conservatively than the $\delta$ and $T$ used for
appearance affinity. This is because the neighbor selection process
for appearance affinity has the additional constraint that the color histograms of the detections
need to look alike. However, for computing spatial affinity, $\tilde{\delta}$ and $\tilde{T}$ are the only
two constraints, thus to avoid connecting incorrect person detections,
they should be set very conservatively.

The loss function which combines the appearance and spatial affinity is as follows:
\begin{equation}
\label{eq:objface}
\begin{gathered}
	\min_\F Tr \left( \F^T (\L + \K) \F \right)  \\
s.t. \hspace{1mm} \textrm{columns of } \F \textrm{ satisfy Equation~\ref{eq:Fdef}}, \;\; \forall i \in \mathcal{Y}, \; \F_i = \Y_i.
\end{gathered}
\end{equation}
Minimizing the loss term will result in a labeling which follows the manifold structure specified by appearance and spatial affinity.
The first term in the constraints specifies that the label assignment matrix $\F$ should be binary and have a single 1 per row.
The second term in the constraints is the face recognition constraint.
Face recognition information is recorded in
$\Y \in \{0, 1\}^{n \times c}$, 
where $\Y_{ij}=1$ if the $i$-th observation belongs to class $j$, i.e. 
the face of observation $i$ is recognized as person $j$.
$\Y_{ij}=0$ if we do not have any label information.
There should only be at most  a single 1 in each row of $\Y$.
$\mathcal{Y} = \left\{ i \mid \exists j \; s.t. \; \Y_{ij} = 1 \right\}$ are all the rows of $\Y$
which have a recognized face.
As face verification is approaching
human-level performance \cite{lu2014surpassing}, it is in most cases reasonable to treat face information as a hard constraint.
Experiments analyzing the effect of face recognition errors on tracking performance are also detailed in Section~\ref{subsec:face_errors}.


\subsection{Spatial Locality Constraint}
\label{sec:slc}

A person cannot be in multiple places at the same time,
and we model this with pairwise person detection constraints.
Given a pair of person detections $(i, j)$, 
if the speed $v^{(ij)}$ required to move from one person detection to the other
is too large, then it is highly unlikely that the pair of person detections will belong to the same person.
We aggregate all the person detection pairs which are highly unlikely to be of the same individual
and encode them in the matrix $\tilde{\S}$, as shown in Equation~\ref{eq:scon}.
\begin{equation}
\label{eq:scon}
\tilde{\S}_{ij} = \left\{ \begin{array}{rl}
		0 & \mbox{ if $v^{(ij)} \le V$ } \\
1 & \mbox{ otherwise } 
\end{array}, \right.
1\le i, j \le n,
\end{equation}
where $V$ is the maximum possible velocity of a moving person.
$\tilde{\S}$ is defined so that 
if none of the person detection velocity constraints were violated,
then $\F_{*j}^T \tilde{\S} \F_{*j} = 0$, where $\F_{*j}$ is the label assignment vector (column vector of $\F$) for the $j$-th person.
We gather this constraint for all individuals and obtain
$Tr(\F^T \tilde{\S} \F) = 0$ if none of the constraints were violated.
The scale of $\tilde{\S}$ is normalized to facilitate the subsequent optimization step.
Let $\D'$ be a diagonal matrix where $\D'_{ii}$ is the sum of row $i$ of $\tilde{\S}$, then we can compute 
the normalized $\S = \D'^{-\frac{1}{2}}\tilde{\S}\D'^{-\frac{1}{2}}$.
The spatial locality constraint is incorporated into our objective function as shown in Equation~\ref{eq:objspar}.
\begin{equation}
\label{eq:objspar}
\begin{gathered}
	\min_\F Tr \left( \F^T (\L + \K) \F \right) \hspace{2mm} s.t. \hspace{2mm} Tr(\F^T \S \F) = 0, \\
  \textrm{ columns of } \F \textrm{ satisfy Equation~\ref{eq:Fdef}}, \; \forall i \in \mathcal{Y}, \F_i = \Y_i.
\end{gathered}
\end{equation}
For simplicity, we do not force two detections from the same frame to not be of the same person.
Nevertheless, this can be easily done by adding additional non-zero elements to $\S$.

Note that the purpose of the affinity-based Laplacian matrix $\L$ and $\K$ are completely opposite of the purpose of $\S$.
$\L$ and $\K$ indicates which observations should be in the same cluster, while $\S$ enforces the fact that two observations cannot be in the same cluster.
Though both $\L$ and $\S$ utilize the same assumption that a person cannot be at multiple places at the same time,
these two matrices have completely different purposes in the loss function.

\subsection{Nonegative Matrix Optimization}
Equation~\ref{eq:objspar} is a combinatorial problem as the values of $\F$ are limited to zeros and ones. 
This is very difficult to solve and certain relaxation is necessary to efficiently solve the objective function.
Therefore, we first relax the form of Equation~\ref{eq:objspar},
and then an iterative projected nonnegative gradient descent procedure is utilized to optimize the relaxed loss function.

The relaxation is motivated as follows.
According to Equation~\ref{eq:FTFI}, the columns of $\F$ are orthogonal
to each other, i.e. $\F^T\F=\J$ is a diagonal matrix.
Also, $\F$ is nonnegative by definition.
According to \cite{NSDR}, if both the orthogonal and nonnegative constraints are satisfied for a matrix,
there will be \emph{at most one non-zero entry} in each row of the matrix. This is still sufficient for
identifying
the class-membership of each observation, i.e. the mutual exclusion constraint still holds
despite the fact that the non-zero entries are no longer exactly 1 but a continuous value.
Therefore, we relax the form of $\F$ by allowing it to take on real values
while still keeping the column orthogonal and nonnegative constraint.
This leads to solving Equation~\ref{eq:objchunkrel}.
\begin{equation}
\label{eq:objchunkrel}
\begin{gathered}
	\min_{\F} Tr \left( \F^T (\L + \K) \F \right) \\
s.t. \hspace{1mm} Tr(\F^T \S \F) = 0, \; \F^T\F = \J, \F \ge 0, \forall i \in \mathcal{Y}, \F_i = \Y_i.
\end{gathered}
\end{equation}
Equation~\ref{eq:objchunkrel} is a constrained quadratic programming problem, in which the mutual exclusion constraint is enforced by $\F^T\F=\J$ and $\F \ge 0$.
One big advantage of this relaxation is that now our method can naturally handle false positive detections,
because $\F$ is now also allowed to have a row where all elements are zeros,
which corresponds to a person detection not being assigned to any class.
This was not possible in the non-relaxed definition of $\F$.
Analysis of robustness against false positives are shown in Section~\ref{subsec:results}.

$\F^T\F = \J$ is still a difficult constraint to optimize. If $\J$ is the identity matrix, then $\F^T \F = \I$ forms
the Stiefel manifold \cite{nmfoja}. Though a few different methods have been proposed
to perform optimization with the orthogonal constraint \cite{nmfoja, nmfding, nmfchoi, nmfalm},
many methods are only applicable to a specific form of the objective function for the optimization process to converge.
Therefore, we instead employ the simple yet effective quadratic penalty method \cite{NSDR,optibook} to optimize the loss function.
The quadratic penalty method incorporates the equality constraints into the loss function by adding a
quadratic constraint violation error for each equality constraint.
The amount of violation is scaled by a weight $\tau$, which gradually increases as more iterations of the optimization are performed,
thus forcing the optimization process to satisfy the constraints.
More details on the convergence properties of the quadratic penalty method can be found in \cite{optibook}.
Therefore, we modify Equation~\ref{eq:objchunkrel} by moving the constraints 
$\F^T\F = \J$ and $Tr \left( \F^T\S\F \right)=0$ 
into the loss function as a penalty term and arrive at the following:
\label{sec:opti}
\begin{equation}
\label{eq:objchunkrelsolve}
\begin{gathered}
	\min_{\F} f\left(\F \right) = \min_{\F} Tr \left( \F^T \left(\L + \K + \tau \S \right) \F \right) + \tau ||\F^T \F - \J||_F^2 \\
	s.t. \hspace{1mm}\F \ge 0, \;\forall i \in \mathcal{Y}, \F_i = \Y_i.
\end{gathered}
\end{equation}
For each $\tau$, we minimize Equation~\ref{eq:objchunkrelsolve} until convergence.
Once converged, $\tau$ is multiplied by a step size $s > 1$ and Equation~\ref{eq:objchunkrelsolve} is minimized again.
Analysis of step size $s$ versus tracking performance is shown in Section~\ref{subsec:results}

To solve for Equation~\ref{eq:objchunkrelsolve} given a fixed $\tau$, 
we perform projected nonnegative gradient descent \cite{lin2007projected},
which iteratively updates the solution at iteration $l$, i.e. $\F^{(l)}$, to $\F^{(l+1)}$:
\begin{equation}
\label{eq:pgd}
\F^{(l+1)} = P \left[ \F^{(l)} - \alpha^{(l)} \nabla \f(\F^{(l)}) \right]
\end{equation}
where the projection function $P$:
\begin{equation}
\label{eq:proj}
P[\F_{ij}] = \left\{ \begin{array}{rl}
		\F_{ij} & \mbox{ if $\F_{ij} > 0$ } \\
0 & \mbox{ otherwise } 
\end{array}, \right.
\end{equation}
is an element-wise function which maps an element back to the feasible region, i.e. in this case a negative number to zero.
The step size $\alpha^{(l)}$ is found in a line search-like fashion, where we search for
an $\alpha^{(l)}$ which provides sufficient decrease in the function value:
\begin{equation}
	\label{eq:condition}
	f(\F^{(l+1)}) - f(\F^{(l)}) \le \sigma Tr\left( \nabla f(\F^{(l)})^T (\F^{(l+1)} - \F^{(l)}) \right).
\end{equation}
Following \cite{lin2007projected}, $\sigma = 0.01$ in our experiments.
The gradient of our loss function $f$ is
\begin{equation}
\label{eq:gradient}
\begin{gathered}
	\nabla f(\F) = 2\left(\L + \K + \tau \S \right) \F + 4\tau \F \left( \F^T\F - \J \right).
\end{gathered}
\end{equation}
Details on convergence guarantees are shown in \cite{lin2007projected}.
To satisfy the face recognition constraints, the values of $\F$ for the rows in $\mathcal{Y}$ are set according to $\Y$
and never updated by the gradient.

The main advantage of projected nonnegative gradient descent over the popular multiplicative updates
for nonnegative matrix factorization \cite{nmf, nmfoja}
is that elements with zero values will have the opportunity to be non-zero in later iterations.
However, for multiplicative updates, zero values will always stay zero.
In our scenario, this means that if $\F^{(l)}_{ij}$ shrinks to $0$ at iteration $l$ in the optimization process,
the decision that ``observation $i$ is not individual $j$'' is final and cannot be changed, which is not ideal.
The projected nonnegative gradient descent method does not have this issue as the updates
are additive and not multiplicative.

$\J$ is a diagonal matrix, where each element on the diagonal $\J_{ii}$ corresponds to the number of 
observations belonging to class $i$, i.e. $m_i$. 
As $m_i$ is unknown beforehand, $m_i$ is estimated by the number of recognized faces belonging to class $i$ plus a constant $\beta$, 
which is proportional to the number of observations $n$. In our experiments we set $\beta = \frac{n}{1000}$.

To initialize our method, we temporarily ignore the mutual exclusion and spatial locality constraint
and only use the manifold and face recognition information to find the initial value $\F^{(0)}$.
$\F^{(0)}$ is obtained by minimizing Equation~\ref{eq:lrga}.
\begin{equation}
\label{eq:lrga}
\small
\min_{\F^{(0)}} Tr \left( (\F^{(0)})^T (\L+\K) \F^{(0)} + (\F^{(0)} - \Y)^T \U (\F^{(0)} - \Y) \right).
\end{equation}
$\U \in \mathbb{R}^{n \times n}$ is a diagonal matrix. 
$\U_{ii} = \infty$ (a large constant) if $i \in \mathcal{Y}$, i.e. the $i$-th observation has a recognized face.
Otherwise $\U_{ii} = 1$.
$\U$ is used to enforce the consistency between prediction results and face recognition label information.
The global optimal solution for Equation~\ref{eq:lrga} is $\F^{(0)}=(\L+\K+\U)^{-1}\U\Y$ \cite{LRGA}.

Once the optimization is complete, we acquire $\F$ which satisfies the mutual exclusion and spatial locality constraint. 
Therefore, trajectories can be computed by simply connecting 
neighboring observations belonging to the same class.
At one time instant, if there are multiple detections assigned to a person, which is common in multi-camera scenarios,
then the weighted average location is computed. The weights are based on the scores in the final solution of $\F$.
A simple filtering process is utilized to remove sporadic predictions.
Algorithm~\ref{alg:algo1} summarizes our tracker.

\begin{algorithm}[t]
	\KwData{Location hypothesis $\p^{(i)}$, $t^{(i)}$, and appearance $\x^{(i)}$, $1 \le i \le n$. Face recognition matrix $\Y\in\{0,1\}^{n\times c}$.}
 \KwResult{Final label assignment matrix $\F$}
 \BlankLine
 Compute Laplacian matrices $\L$, $\K$ \tcp*[r]{\textrm{Sec.~\ref{sec:laplacian}}} 
 Compute spatial locality matrix $\S$ \tcp*[r]{\textrm{Sec.~\ref{sec:slc}}} 
 Compute diagonal matrix $\J$ \tcp*[r]{\textrm{Sec.~\ref{sec:opti}}} 
 Compute diagonal matrix $\U$ from $\Y$ \tcp*[r]{\textrm{Sec.~\ref{sec:opti}}} 
 Initialize $\F^{(0)}$ with Equation~\ref{eq:lrga} \;
 $l \leftarrow 0$ \tcp*[r]{\textrm{iteration count}}
 $\tau \leftarrow 10^{-4}$ \tcp*[r]{\textrm{initial penalty}}
 \Repeat(\tcp*[f]{\textrm{Solve for Equation~\ref{eq:objchunkrelsolve} with penalty method}}){$\tau \ge 10^{11}$}{
  $\tau \leftarrow \tau \times s$ \tcp*[r]{\textrm{gradually increase penalty $\tau$}}
  \Repeat(\tcp*[f]{\textrm{projected gradient descent}}){convergence}{
   Compute $\F^{(l+1)}$ from $\F^{(l)}$ with Equation~\ref{eq:pgd}\;
   $l \leftarrow l+1$\;
  }
 }
 \Return{$\F^{(l)}$}
 \vspace{2mm}
 \caption{Main steps in proposed tracking algorithm.}
 \label{alg:algo1}
\end{algorithm}

\section{Experiments}
\label{sec:results}

We present experiments on tracking
followed by video summarization experiments based on our long-term tracking output.

\subsection{Tracking}

\subsubsection{Data Sets}
As we are interested in evaluating identity-aware tracking,
we focused on sequences where identity information such as face recognition was available.
Therefore, many popular tracking sequences such as the PETS 2009 sequences \cite{PETS09}, Virat \cite{virat}, TRECVID 2008 \cite{SED} and Town Centre \cite{benfold2011stable}
were not applicable as the faces in these sequences were too small to be recognized and no other identity information could be extracted.
Basketball related sequences \cite{TMCNF,vondrick2013efficiently} were not used as some 
manual effort is required to have an accurate OCR of jersey numbers \cite{TMCNF}.
The following four data sets were utilized in our experiments.

\noindent
\textbf{\emph{terrace1}}: The 4 camera \emph{terrace1} \cite{POM} data set has 9 people walking around in a 7.5m by 11m area for 3 minutes 20 seconds.
The scene is very crowded, thus putting the spatial locality constraint to test.
The POM grid we computed had width and height of 25 centimeters per cell.
Person detections were extracted at every frame.
As the resolution of the video is low, one person did not have a recognizable face.
For the sake of performing identity-aware tracking on this dataset,
we manually added two identity annotations for each individual at the start and end of the person's trajectory
to guarantee that each individual had identity labels.
None of the trackers utilized the fact that these two additional annotations were the start and end of a trajectory.
In total, there were 794 identity labels out of 57,202 person detections.

\noindent
\textbf{\emph{Caremedia 6m}}: The 15 camera \emph{Caremedia 6m}
\cite{Caremedia, Marauders} data set has 13 individuals performing daily activities in a nursing
home for 6 minutes 17 seconds.
Manual annotations were provided every second and interpolated to every frame.
The data set records activities in a nursing home where
staff maintain the nursing home and assist residents throughout the day.
As the data set covers a larger area and is also longer than \emph{terrace1}, we ran into memory
issues for trackers which take POM as input when our cell size was 25 centimeters.
Therefore, the POM grid we computed had width and height of 40 centimeters per cell.
Person detections were extracted at every sixth frame.
In total, there were 2,808 recognized faces and 12,129 person detections.
Though on average there was a face for every 4 detections, but
recognized faces were usually found in clusters and not evenly spread out over time. 
So there were still periods of time
when no faces were recognized.

\noindent
\textbf{\emph{Caremedia 8h}}:
The 15 camera \emph{Caremedia 8h}
data set is a newly annotated data set which 
has 49 individuals performing daily activities in the same nursing home as \emph{Caremedia 6m}.
The sequence is 7 hours 45 minutes long, which is 116.25 hours of video in total.
Ground truth was annotated every minute.
Person detections were extracted at every sixth frame.
In total, there were 70,994 recognized faces and 402,833 person detections.

\noindent
\textbf{\emph{Caremedia 23d}}:
The 15 camera \emph{Caremedia 23d}
data set is a newly annotated data set which consists of nursing home recordings spanning over 23 days.
Recordings at night were not processed as there was not much activity at night.
In total, 4,935 hours of video were processed.
To the best of our knowledge, this is the longest sequence to date to be utilized for multi-object tracking experiments.
\emph{Caremedia 23d}
has 65 unique individuals.
Ground truth was annotated every 30 minutes.
Person detections were extracted at every sixth frame.
In total, there were 3.1 million recognized faces and 17.8 million person detections. 

\subsubsection{Baselines}
\label{subsec:baselines}
We compared our method with three identity-aware tracking baselines.
As discussed in the Related Work section (Section~\ref{sec:related}), it is non-trivial to modify a non-identity-aware tracker
to incorporate identity information.
Therefore, other trackers which did not have the ability to incorporate identity information
were not compared.

\noindent
\textbf{Multi-Commodity Network Flow (MCNF)}: The MCNF tracker \cite{TMCNF} can be viewed as the K-Shortest-Path tracker (KSP, \cite{KSP})
with identity aware capabilities.
The KSP is a network flow-based method
that utilizes POM localization information. 
Based on POM, the algorithm will find the $K$ shortest paths,
which correspond to the $K$ most likely trajectories in the scene.
MCNF further duplicates the graph in KSP for every different identity group in the scene.
The problem is solved with linear programming plus an additional step of rounding non-integral values.
We reimplemented the MCNF algorithm.
The graph was duplicated $c$ times to reflect the $c$ unique individuals.
Gurobi \cite{gurobi} was used as our linear program solver.
Global appearance templates were computed from person detections which had recognized faces.
The source code of POM and KSP were from the authors \cite{POM, KSP}.
This setting is referred to as \emph{MCNF w/ POM}. 
The base cost of generating a trajectory, which is a parameter that controls the minimum length of the generated tracks, is set to -185
for all \emph{MCNF w/ POM} experiments.
For the two Caremedia data sets, we also took the person detection (PD) output and generated POM-like localizations which
were also provided to MCNF. The localizations were generated by 
aggregating all person detections falling into each discretized grid cell at each time instant.
This setting is referred to as \emph{MCNF w/ PD}.
For all \emph{MCNF w/ PD} experiments, the grid size is 40 centimeters, the base cost of generating a trajectory is -60, and
detections were aggregated over a time span of 6 frames to prevent broken trajectories.
For the \emph{Caremedia 8h} and \emph{Caremedia 23d} set, the Gurobi solver was run in 12,000 frame batches to avoid memory issues.

\noindent
\textbf{Lagrangian Relaxation (LR)}: \cite{dehghan2015target} utilized LR to impose mutual exclusion constraints for identity-aware tracking
in a network flow framework very similar to MCNF, where each identity has their own identity specific edges.
To fairly compare different data association methods, our LR-based tracker utilized the same appearance information used by all our other trackers,
thus the structured learning and densely sampled windows proposed in \cite{dehghan2015target} were not used.
Specifically, LR uses the same POM-like input and network as MCNF.

\noindent
\textbf{Non-Negative Discretization (NND)}: The Non-Negative Discretization tracker \cite{Marauders} is a primitive version of our proposed tracker.
The three main differences are: 1) NND does not have the spatial locality constraint, 2) an extra Viterbi
trajectory formulation step, which requires the start and end of trajectories, was necessary, and 
3) a multiplicative update was used to perform non-negative matrix factorization.
Start and end locations of trajectories are often unavailable in real world scenarios.
Therefore, no start and end locations were provided to NND in our experiments,
and the final trajectories of NND were formed with the same method used by our proposed tracker.
NND utilizes \cite{LRGA} to build the manifold, but internal experiments have shown that utilizing the method in \cite{LRGA}
to build the Laplacian matrix achieves similar tracking performance compared to the standard method \cite{ng2002spectral,belkin2003laplacian}.
Therefore, to fairly compare the two data association methods, we utilized the same Laplacian matrix computation method for NND and our method.
Also the spatial affinity term $\K$ was not used in the originally proposed NND, but for fairness we added the $\K$ term to NND.

\subsubsection{Implementation Details}
We utilized the person detection model from \cite{persondetect,voc-release5} for person detection.
Color histograms for the person detection were computed the same way as in \cite{Marauders}.
We used HSV color histograms as done in \cite{Okuma04aboosted}.
We split the bounding box horizontally into regions and computed
the color histogram for each region similar to the spatial pyramid matching technique \cite{SPM}. 
Given $L$ layers, we have $2^{L} - 1$ partitions for each template. 
$L$ was 3 in our experiments.
Since the person detector only detects upright people,
tracking was not performed on sitting people or residents in wheelchairs.
Background subtraction for POM was performed with \cite{stauffer1999adaptive}.
Face information is acquired from the PittPatt 
software\footnote{Pittsburgh Pattern Recognition (http://www.pittpatt.com)},
which can recognize a face when a person is close enough to the camera.
We acquired the gallery by clustering the recognized faces and then manually
assigning identities to each cluster.

For our proposed method, the parameters for all four data sets were as follows.
The number of nearest neighbors used for appearance-based manifold construction was $k=25$.
The window to search for appearance-based nearest neighbors was $T=8$ seconds.
The color histogram threshold $\gamma=0.85$.
The maximum localization error was $\delta=$ 125 cm.
For modeling spatial affinity, $\tilde{\delta}$ was 20 cm, and $\tilde{T}$ was 6 frames.
When computing the spatial locality constraint matrix $\S$, 
we only looked for conflicting observations which were less than 6 frames apart to retain sparse $\S$.
The above parameters were also used for NND.
For the optimization step, the initial value of $\tau=2\times 10^{-4}$, and the final value was $\tau=10^{11}$.
The step size for updating $\tau$, i.e. $\tau \leftarrow \tau \times s$, is $s = 2$.


\subsubsection{Evaluation Metrics}
Identity-aware tracking can be evaluated from a multi-object tracking point of view and a classification point of view.
From the tracking point of view,
the most commonly used multi-object tracking metric is 
\emph{Multiple Object Tracking Accuracy} (MOTA\footnote{Code modified from http://www.micc.unifi.it/lisanti/source-code/.}) \cite{CLEAR,MOTA}.
Following the evaluation method used in 
\cite{dctracker,Marauders}, the association between the tracking results and the ground truth
is computed in 3D with a hit/miss threshold of 1 meter.
MOTA takes into account the number of true positives (TP), 
false positives (FP), missed detections (false negatives, FN) and identity switches (ID-S).
Following the setting in \cite{TMCNF}
\footnote{There are two common transformation functions (denoted as $c_s()$ in \cite{MOTA}) for the identity-switch term, 
	either $\log_{10}$ \cite{MOTA,TMCNF} 
	or the identity function \cite{CLEAR}. We have selected the former as this is what was used in MCNF, which is one of our baselines.}
MOTA is computed as follows:
$1 - \frac{\textrm{\# FP} + \textrm{\# FN} + \log_{10}(\textrm{\# ID-S})}{\textrm{\# ground truth}}$.

However, the TP count in MOTA does not take into account the identity of a person,
which is unreasonable for identity aware tracking.
Therefore, we compute identity-aware true positives (I-TP), which means that a detection is only a true positive
if 1) it is less than 1 meter from the ground-truth and 2) the identities match.
Similarly, we can compute I-FP and I-FN, which enables us to compute classification-based metrics such as
micro-precision ($\textrm{MP} = \frac{\textrm{\# I-TP}}{\textrm{\# I-TP + \# I-FP}}$), 
micro-recall ($\textrm{MR} = \frac{\textrm{\# I-TP}}{\textrm{\# I-TP + \# I-FN}}$) and a comprehensive 
micro-F1 ($\frac{\textrm{2 $\times$ MP $\times$ MR}}{\textrm{MP+MR}}$) 
for each tracker.
The \emph{micro}-based performance evaluation takes into account the length (in terms of time) of each person's trajectory, 
so a person who appears more often has larger influence to the final scores.

\subsubsection{Tracking Results}
\label{subsec:results}

\begin{table}[tp]
\begin{subtable}{\linewidth}\centering
	\scriptsize
{
	\begin{tabular}{ | @{\hspace{1pt}}C{1.75cm}@{\hspace{1pt}} || @{\hspace{1pt}}C{1.0cm}@{\hspace{1pt}} | @{\hspace{1pt}}C{0.8cm}@{\hspace{1pt}} | @{\hspace{1pt}}C{0.8cm}@{\hspace{1pt}} || @{\hspace{1pt}}C{0.7cm}@{\hspace{1pt}} | @{\hspace{1pt}}C{0.7cm}@{\hspace{1pt}} | @{\hspace{1pt}}C{0.7cm}@{\hspace{1pt}} | @{\hspace{1pt}}C{0.6cm}@{\hspace{1pt}} | @{\hspace{1pt}}C{0.8cm}@{\hspace{1pt}} | }
		\hline
		Method		& Micro-Precision & Micro-Recall & \textbf{Micro-F1} & TP & FN & FP & ID-S & \textbf{MOTA}  \\ \hline 
		\emph{Face only} & 0.493 & 0.018 & 0.035 & 646 & 24708 & 284 & 5 & 0.014 \\ \hline
		\emph{MCNF w/ POM} & 0.593 & 0.532 & 0.561 & 21864 & 3298 & 644 & 197 & \textbf{0.844} \\ \hline 
		\emph{LR w/ POM} & 0.609 & 0.478 & 0.535 & 19216 & 5996 & 521 & 147 & 0.743 \\ \hline
		\emph{NND}  & 0.613 & 0.238 & 0.343 & 8035 & 17267 & 1771 & 57 & 0.249\\ \hline 
		\emph{Ours w/o SLC}  & 0.704 & 0.346 & 0.464 & 10642 & 14655 & 1745 & 62 & 0.353 \\ \hline
		\emph{Ours}  & 0.692 & 0.635 & \textbf{0.663} & 21370 & 3873 & 1783 & 116 & 0.777 \\ \hline
	\end{tabular} 
}
\vspace{-2mm}
\caption{Tracking performance on \emph{terrace1} sequence.}
\label{tab:trackingterrace1}
\end{subtable}
\begin{subtable}{\linewidth}\centering
{
	\scriptsize
	\begin{tabular}{ | @{\hspace{1pt}}C{1.75cm}@{\hspace{1pt}} || @{\hspace{1pt}}C{1.0cm}@{\hspace{1pt}} | @{\hspace{1pt}}C{0.8cm}@{\hspace{1pt}} | @{\hspace{1pt}}C{0.8cm}@{\hspace{1pt}} || @{\hspace{1pt}}C{0.7cm}@{\hspace{1pt}} | @{\hspace{1pt}}C{0.7cm}@{\hspace{1pt}} | @{\hspace{1pt}}C{0.7cm}@{\hspace{1pt}} | @{\hspace{1pt}}C{0.6cm}@{\hspace{1pt}} | @{\hspace{1pt}}C{0.8cm}@{\hspace{1pt}} | }
		\hline
		Method		& Micro-Precision & Micro-Recall & \textbf{Micro-F1} & TP & FN & FP & ID-S & \textbf{MOTA}  \\ \hline 
		\emph{Face only} & 0.942 & 0.362 & 0.523 & 12369 & 21641 & 727 & 9 & 0.342 \\ \hline
		\emph{MCNF w/ POM} & 0.117 & 0.238 & 0.157 & 23493 & 9769 & 44452 & 757 & -0.594 \\ \hline
		\emph{MCNF w/ PD} & 0.746 & 0.578 & 0.652 & 19941 & 13749 & 5927 & 329 & 0.422 \\ \hline 
		\emph{LR w/ PD} & 0.802 & 0.565 & 0.663 & 19415 &  14408 & 4203 & 196 & 0.453 \\ \hline 
		\emph{NND} & 0.861 & 0.726 & 0.787 & 25628 & 8364 & 3100 & 27 & 0.663 \\ \hline 
		\emph{Ours w/o SLC} &  0.869 & 0.726 & 0.791 & 25578 & 8408 & 3080 & 33 & 0.662\\ \hline
		\emph{Ours} 	& 0.865 & 0.755 & \textbf{0.807} & 26384 & 7576 & 3537 & 59 & \textbf{0.673}  \\ \hline
	\end{tabular} 
}
\vspace{-2mm}
\caption{Tracking performance on \emph{Caremedia 6m} sequence.}
\label{tab:tracking617}
\end{subtable}
\begin{subtable}{\linewidth}\centering
{
	\scriptsize
	\begin{tabular}{ | @{\hspace{1pt}}C{1.75cm}@{\hspace{1pt}} || @{\hspace{1pt}}C{1.0cm}@{\hspace{1pt}} | @{\hspace{1pt}}C{0.8cm}@{\hspace{1pt}} | @{\hspace{1pt}}C{0.8cm}@{\hspace{1pt}} || @{\hspace{1pt}}C{0.7cm}@{\hspace{1pt}} | @{\hspace{1pt}}C{0.7cm}@{\hspace{1pt}} | @{\hspace{1pt}}C{0.7cm}@{\hspace{1pt}} | @{\hspace{1pt}}C{0.6cm}@{\hspace{1pt}} | @{\hspace{1pt}}C{0.8cm}@{\hspace{1pt}} | }
		\hline
		Method		& Micro-Precision & Micro-Recall & \textbf{Micro-F1} & TP & FN & FP & ID-S & \textbf{MOTA}  \\ \hline 
		\emph{Face only} & 0.858 & 0.256 & 0.394 & 164 & 471 & 19 & 2 & 0.230 \\ \hline
		\emph{MCNF w/ PD} & 0.743 & 0.418 & 0.535 & 265 & 347 & 71 & 25 & 0.342 \\ \hline 
		\emph{LR w/ PD} & 0.787 & 0.405 & 0.535 & 261 & 360 & 52 & 16 & 0.351\\ \hline 
		\emph{NND} & 0.588 & 0.505 & 0.543 & 314 & 281 & 174 & 42 & 0.283\\  \hline 
		\emph{Ours w/o SLC} & 0.638 & 0.549 & 0.590 & 349 & 257 & 151 & 31 & 0.357 \\ \hline
		\emph{Ours} & 0.648 & 0.571 & \textbf{0.607} & 370 & 241 & 149 & 26 & \textbf{0.386} \\ \hline
	\end{tabular} 
}
\vspace{-2mm}
\caption{Tracking performance on \emph{Caremedia 8h} sequence.}
	\label{tab:tracking745}
\end{subtable}
\begin{subtable}{\linewidth}\centering
{
	\scriptsize
	\begin{tabular}{ | @{\hspace{1pt}}C{1.75cm}@{\hspace{1pt}} || @{\hspace{1pt}}C{1.0cm}@{\hspace{1pt}} | @{\hspace{1pt}}C{0.8cm}@{\hspace{1pt}} | @{\hspace{1pt}}C{0.8cm}@{\hspace{1pt}} || @{\hspace{1pt}}C{0.7cm}@{\hspace{1pt}} | @{\hspace{1pt}}C{0.7cm}@{\hspace{1pt}} | @{\hspace{1pt}}C{0.7cm}@{\hspace{1pt}} | @{\hspace{1pt}}C{0.6cm}@{\hspace{1pt}} | @{\hspace{1pt}}C{0.8cm}@{\hspace{1pt}} | }
		\hline
		Method		& Micro-Precision & Micro-Recall & \textbf{Micro-F1} & TP & FN & FP & ID-S & \textbf{MOTA}  \\ \hline 
		\emph{Face only} & 0.819 & 0.199 & 0.154 & 125 & 512 & 28 & 2 & 0.154 \\ \hline
		\emph{MCNF w/ PD} & 0.712 & 0.355 & 0.474 & 205 & 412 & 92 & 22 & 0.209 \\ \hline
		\emph{LR w/ PD} & 0.663 & 0.357 & 0.464 & 215 & 411 & 116 & 13 & 0.174 \\ \hline
        \emph{Ours} & 0.698 & 0.532 & \textbf{0.604} & 326 & 299 & 147 & 14 & \textbf{0.300} \\ \hline
	\end{tabular} 
}
\vspace{-2mm}
\caption{Tracking performance on \emph{Caremedia 23d} sequence.}
	\label{tab:tracking23d}
\end{subtable}
\caption{Tracking performance on 4 tracking sequences. 
POM: Probabilistic Occupancy Map proposed in \cite{POM} as input.
PD: Person detection as input.
SLC: Spatial locality constraint.
``w/'' and ``w/o'' are shorthand for ``with'' and ``without'' respectively.
We did not run the \emph{MCNF w/ POM} on the longer Caremedia sequences as it was already performing poorly on \emph{Caremedia 6m}.
}
\label{tab:tracking}
\end{table}

\begin{figure}[tp]
\centering
\includegraphics[scale=0.22]{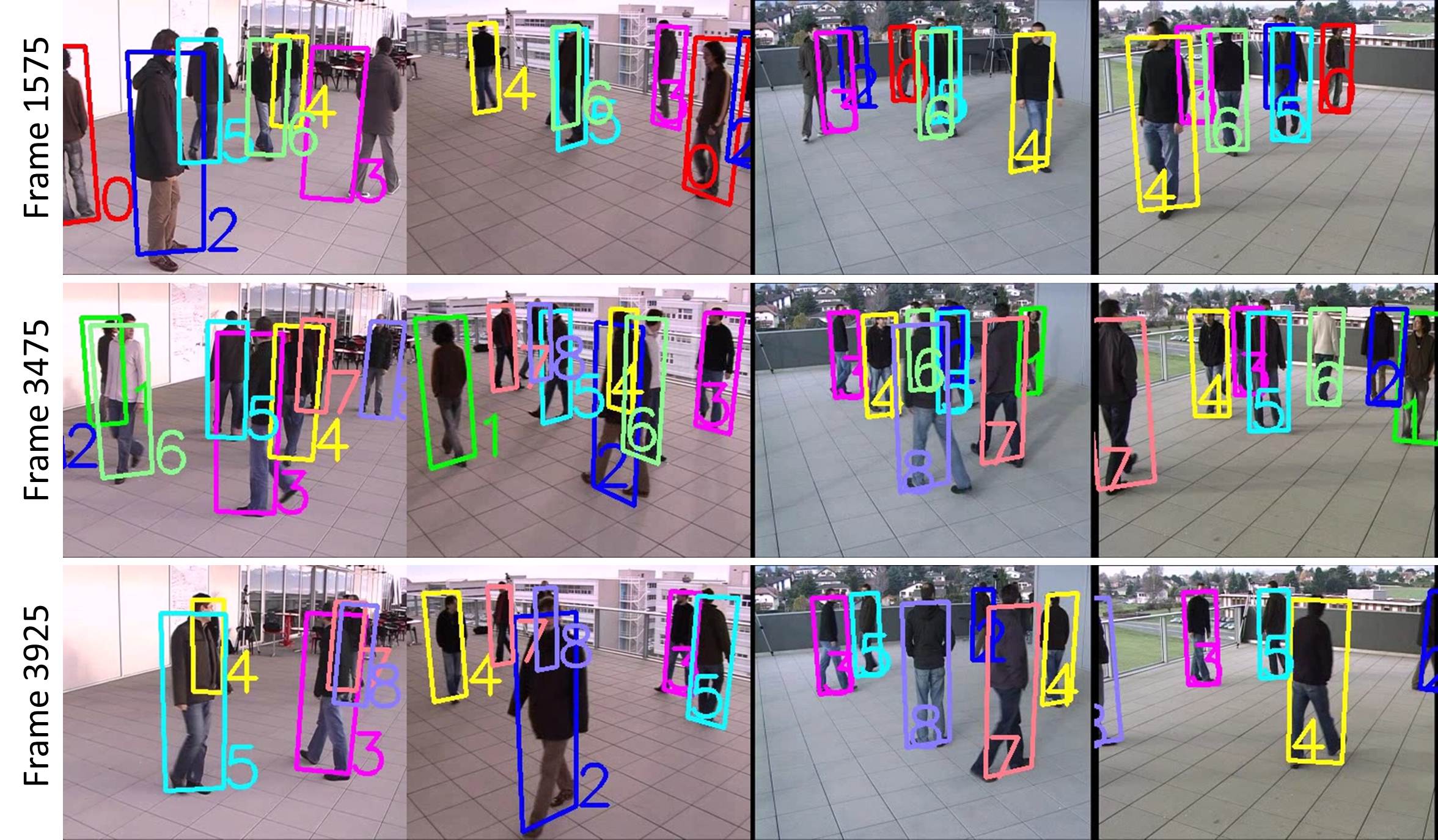}
\caption{
	Snapshots of tracking results from the 4 camera \emph{terrace1} sequence.
}
\label{fig:epflsnap}
\end{figure}

\begin{figure*}[tp]
\centering
\begin{subfigure}{0.48\textwidth}
\centering
\includegraphics[scale=0.22]{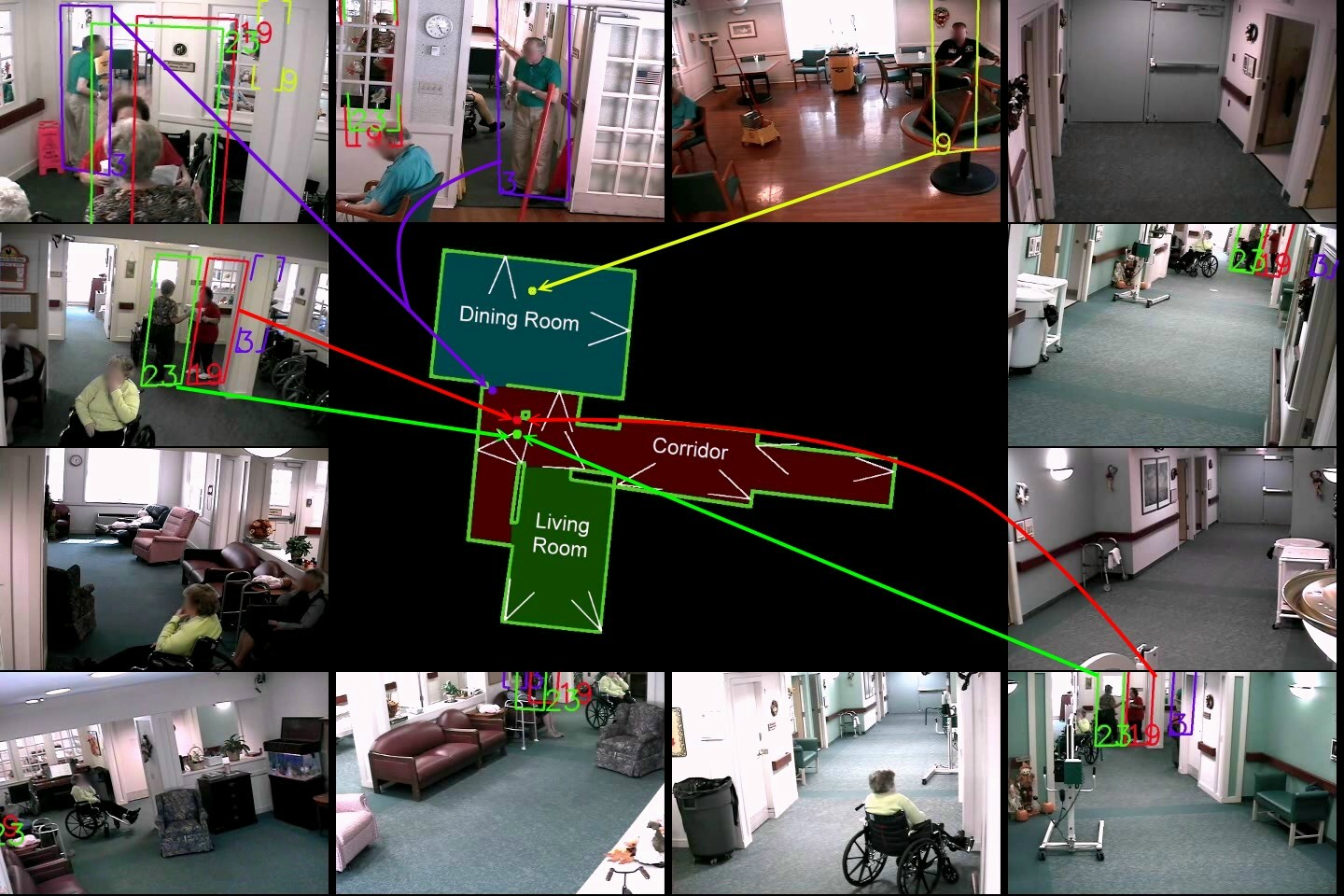}\label{fig:snap1}
\end{subfigure}
\begin{subfigure}{0.48\textwidth}
\hspace{1mm}
\centering
\includegraphics[scale=0.22]{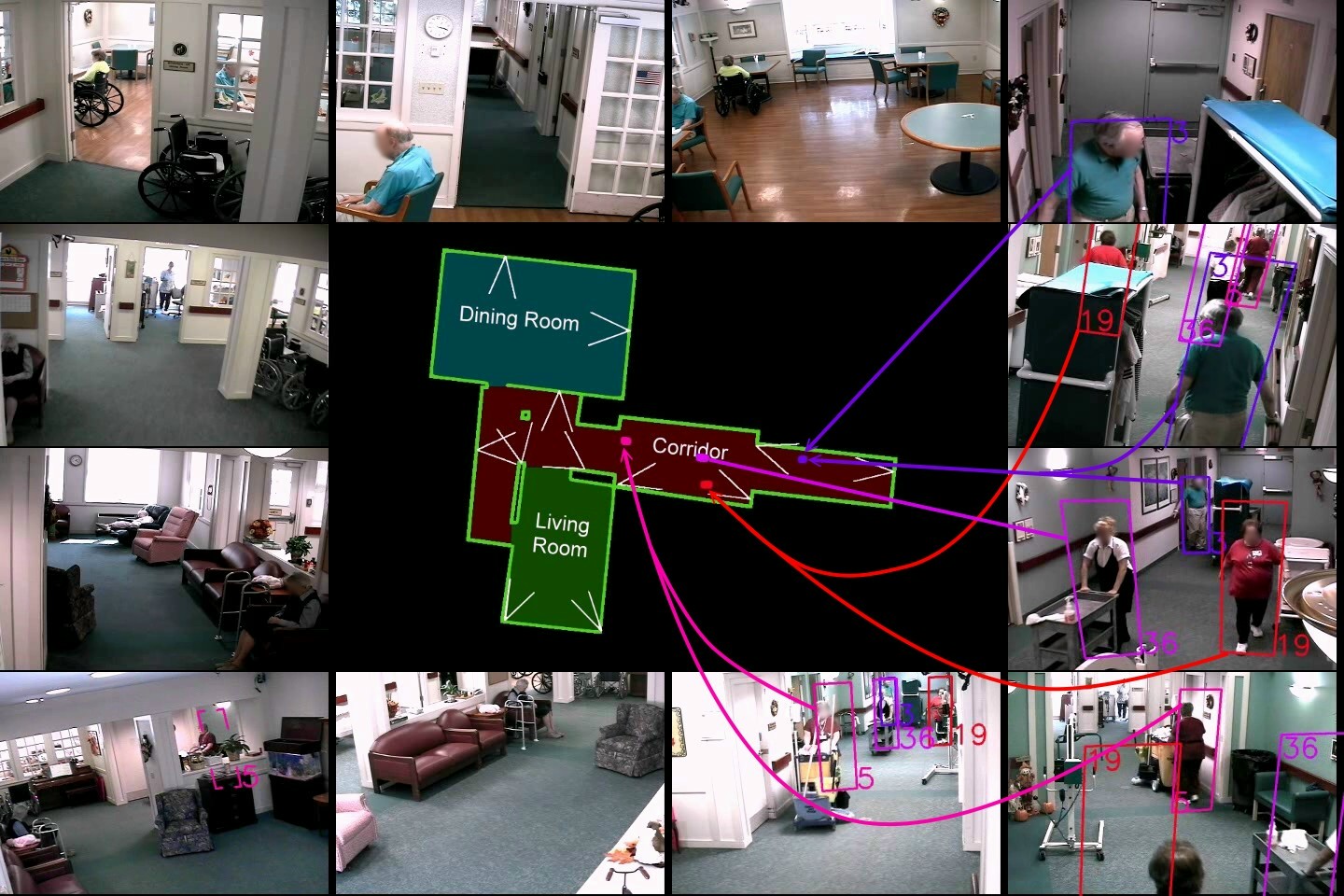}\label{fig:snap2}
\end{subfigure}
\caption{
Snapshots of tracking results from \emph{Caremedia 8h} data set.
To increase readability, not all arrows are drawn and only 12 out of 15 cameras are shown.
}
\label{fig:caresnap}
\end{figure*}

Tracking results for the four data sets are shown in Table~\ref{tab:tracking}.
We achieve the best performance
in F1-scores across all four data sets. This means that our tracker
can not only track a person well, but can also accurately identify the individual.
Figure~\ref{fig:epflsnap} and Figure~\ref{fig:caresnap} show qualitative examples of our tracking result.

The importance of the spatial locality constraint (SLC) is also shown clearly in Table~\ref{tab:trackingterrace1}.
Without the spatial locality constraint in the optimization step (\emph{NND} and \emph{Ours w/o SLC}), 
performance degraded significantly in the very crowded \emph{terrace1} sequence as
the final result may show a person being
at multiple places at the same time, thus hijacking the person detections of other individuals.
For the \emph{Caremedia} sequences, the SLC does not make a big difference, because 1) the scene is not so crowded and 2) the appearance
of each individual is more distinct, thus relying only on the appearance feature can already achieve good performance.

The performance of \emph{Face only} clearly shows the contribution of face recognition and tracking.
For the Caremedia related sequences, face recognition could already achieve certain performance,
but our tracker further improved F1 by at least 20\% absolute.
For \emph{terrace1}, there were very limited faces, and we were able to increase F1 by 60\% absolute.

We also analyzed the robustness of our algorithm against false positives.
The person detections on \emph{Caremedia 6m} had around 13\% false positive rate.
Manual verification
showed that for the person detections that were assigned a label by our tracker,
only 0.1\% were false positive detections.
This means that $\frac{12.9\%}{13\%}=99\%$ of the false positives were filtered out by our algorithm,
thus demonstrating the robustness of our method against false positives.

\begin{figure}[tp]
\centering
\includegraphics[scale=0.4]{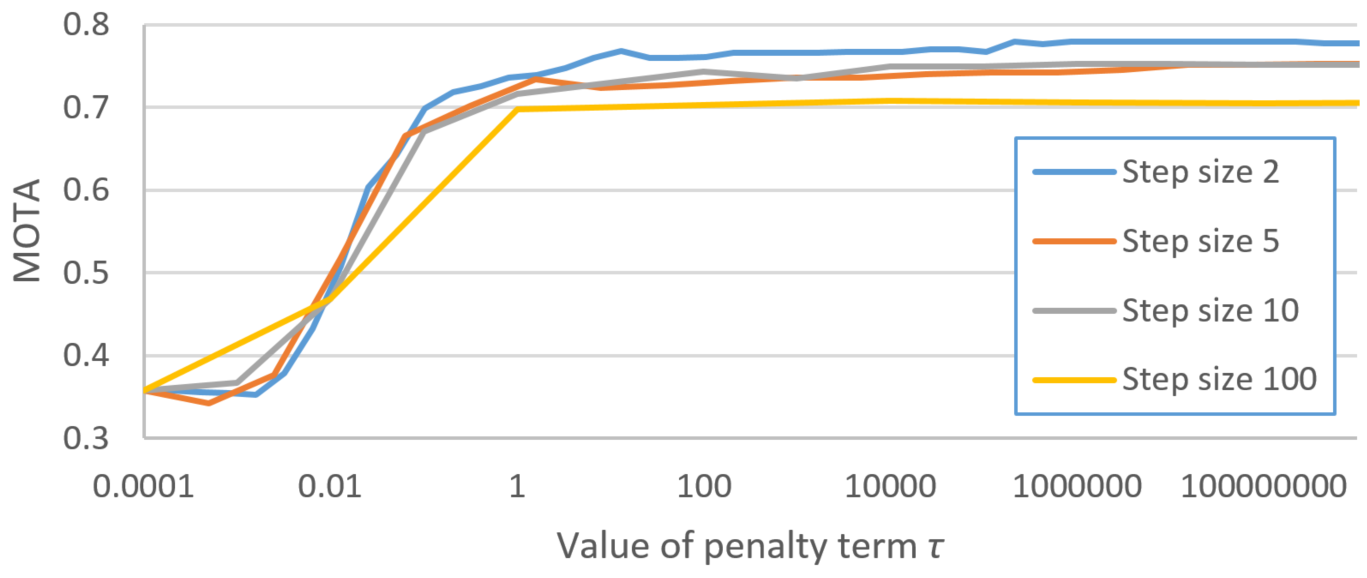}
\caption{
	Performance of \emph{terrace1} MOTA under different penalty term $\tau$ and step size $s$. Best viewed in color.
}
\label{fig:penalty}
\end{figure}

Figure~\ref{fig:penalty} demonstrates the effect of using different step size $s$ when increasing the penalty
term $\tau$, which is utilized to 
enforce the mutual exclusion and spatial locality constraints.
The initialization of our optimization process (Equation~\ref{eq:lrga}) does not enforce the two constraints,
which lead to a MOTA of 0.358 when $\tau=0$. As $\tau$ increases, MOTA gradually increased to 0.777, 
which demonstrates 1) the constraints were very important
and 2) the quadratic penalty term utilized effectively enforced these constraints.
Also, if the penalty term $\tau$ was increased too quickly, i.e. $s$ is large, 
then tracking performance drops. This is reasonable as the optimization process is prone to
getting stuck in a bad local minimum when the solution acquired from the previous $\tau$
is not a good initialization for the next $\tau$.

The MCNF tracker is also a very strong baseline.
For \emph{terrace1}, KSP and consequently MCNF achieved very good MOTA results with POM person localization.
MCNF was slightly worse than KSP on MOTA scores because 1) though MCNF is initialized by KSP, MCNF is no longer solving a problem
with a global optimal solution and 2) MCNF is not directly optimizing for MOTA.
However, for the \emph{Caremedia 6m} sequence, \emph{MCNF with POM} performance was poor
because POM created many false positives in
the complex indoor nursing home environment.
This is due to non-ideal camera coverage that caused ambiguities in POM localization. 
Nevertheless, if the person detections used in our method was provided to MCNF (\emph{MCNF with PD}),
then MCNF performs reasonably well.

For \emph{Caremedia 23d}, 
our best tracker can locate a person 53.2\% of the time with 69.8\% precision,
i.e. in a 23 day time span, we can find a person more than 50\% of the time with 70\% accuracy.
These results are encouraging, as the tracking output with such performance already has the potential to be utilized
by other tasks, such as the experiments performed in Section~\ref{sec:event} on surveillance video summarization.

\subsubsection{Discussion - Advantages of Tracker}
\label{subsec:adv}
The key advantages of our tracker are as follows:

\noindent
\textbf{Face recognition output is integrated into the framework:}
Face recognition serves as a natural way to automatically assign identities to trajectories and also
reinitialize trajectories in long-term tracking scenarios,
where manual intervention is prohibitively costly.
Also, face recognition is not affected when the same person wear different clothing in recordings over multiple days.

\noindent
\textbf{Naturally handle appearance changes:}
In our tracker,
the appearance templates of the tracked target are implicitly encoded in the manifold structure we learn. 
Therefore, if the appearance of a tracked object changes smoothly along a manifold, our
algorithm can model the change.
No threshold is required to decide when to adaptively update the appearance model.
If there is a drastic change in appearance for a tracked object, then the appearance manifold will highly likely be broken.
However, the spatial affinity term could still link up the manifold.

\noindent
\textbf{Take into account appearance from multiple neighbors:}
Our tracker takes into account appearance information from multiple neighboring points,
which enables us to have a more stable model of appearance.
Linear programming and network flow-based methods can only either have a global appearance model or
model appearance similarity only over the previous and next detection in the trajectory.

\noindent
\textbf{Handle multiple detections per frame for one individual:}
In multi-camera scenes, it is common that at one time instant, multiple detections from different cameras correspond to the same physical person.
This may be difficult to deal with
for single-camera multi-object trackers based on network flow \cite{zhang2008global,pirsiavash2011globally},
because the spatial locality constraint for these methods are
enforced based on the assumption that each individual can only be assigned a single person detection per frame.
Therefore, multi-camera network flow-based methods such as \cite{KSP, TMCNF} utilize a two-step process where the POM is first used to aggregate evidences from
multiple cameras before performing data association.
Our formulation of the spatial locality constraint, which is based on the velocity to travel between two detections being below a threshold,
can be viewed as a generalization to the aforementioned assumption,
and this enables us to have localization and data association in a single optimization framework.


\noindent
\textbf{No discretization of the space required in multi-camera scenarios:}
Previous multi-camera network flow methods \cite{KSP, TMCNF} require discretization of the tracking space in multi-camera scenarios
to make the computation feasible.
Finer grids run into memory issues when the tracking sequence is long and covers a wide area,
and coarser grids run the risk of losing precision.
However, our tracker works directly on person detections, and discretization is not necessary.

\subsubsection{Discussion - Limitations of Tracker}
There are also limitations to our tracker.

\noindent
\textbf{Assumes at least one face recognition per trajectory:} 
If there is a trajectory where no faces were observed and recognized, then our tracker will completely ignore this trajectory, which
is acceptable if we are only interested in identity-aware tracking.
Otherwise,
one potential solution is to find clusters of unassigned person detections and assign pseudo-identities to them
to recover the trajectories.

\noindent
\textbf{Only bounded velocity model employed:}
To employ the more sophisticated constant velocity model,
we could use pairs of points as the unit of location hypotheses, but this may generate significantly more
location hypotheses than the current approach.

\noindent
\textbf{Assumes all cameras are calibrated:}
To combine person detections from different camera views, we utilize camera calibration parameters to map all person detections into a global coordinate system.

\noindent
\textbf{Face recognition gallery required beforehand:} 
In order to track persons-of-interest, we require the gallery beforehand.
This is the only manual step in our whole system,
which could be alleviated by face clustering. Face clustering enables humans to efficiently assign identities to each cluster.
Also, in a nursing home setting, the people-of-interest are fixed,
thus this is a one-time effort which could be used for weeks or even months of recordings.

\begin{figure}[tp]
\centering
\includegraphics[scale=0.3]{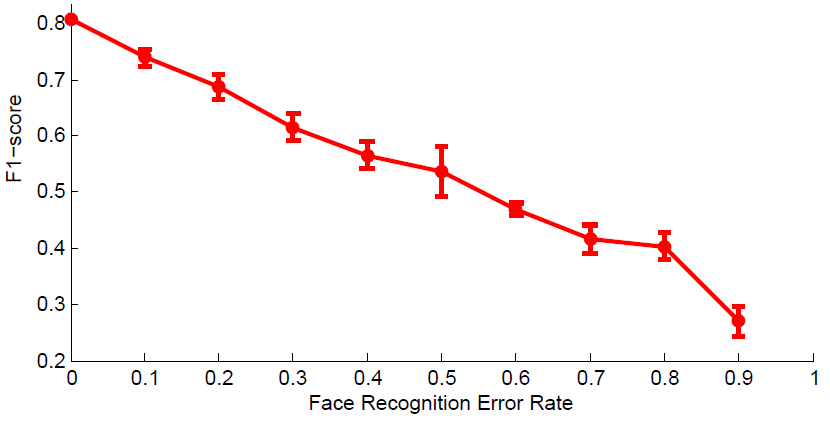}
\caption{
	\emph{Caremedia 6m} tracking performance under varying face recognition error rate.
}
\label{fig:face_error}
\end{figure}

\noindent
\textbf{Assumes perfect face recognition:} \label{subsec:face_errors}
The current framework assumes perfect face recognition, which may not be applicable in all scenarios.
We analyzed the effect of face recognition accuracy on tracking performance.
We generated face recognition errors by randomly corrupting face recognition results in the \emph{Caremedia 6m} set.
The error rates range from 10\% to 90\%.
The experiment was repeated 3 times per error rate, and the results with the 95\% confidence intervals are shown in Figure~\ref{fig:face_error}.
Results show that the general trend is a 20\% increase in face recognition error will cause around 10\% drop in tracking F1-score.

\subsubsection{Timing Analysis}
The whole tracking system includes person detection, face recognition, color histogram
extraction and data association.
The person detector we utililized \cite{persondetect,voc-release5} ran at 40 times real-time.
However, recently proposed real-time person detectors \cite{DPM30Hz}
will enable us to run person detection at 1 time real-time.
The rest of the pipeline runs at around 3 times real-time on a single core, and the pipeline can be easily parallelized to run faster than real-time.
The data association part, which is our main focus,
runs at around $\frac{1}{40}$ times real-time.

\subsection{Visual Diary Generation}
\label{sec:event}

To demonstrate the usefulness of our tracking output, video summarization experiments were performed.
We propose to summarize surveillance video using visual diaries, specifically in the context of monitoring elderly residents
in a nursing home.
Visual diary generation for elderly
nursing home residents could enable doctors and staff to quickly understand
the activities of a senior person throughout the day to facilitate
the diagnosis of the elderly person's state of health.
The visual diary for a specific person consists of two parts as shown in Figure~\ref{fig:diary11}: 1) snippets which contain
snapshots and textual descriptions of activities-of-interest performed
by the person, and 2) activity-related statistics accumulated over the whole day.
The textual descriptions of the detected events enables efficient indexing of what a person did at different times.
The statistics for the activities detected can be accumulated over many days to discover long-term patterns.

We propose to generate visual diaries with a summarization-by-tracking framework.
Using the trajectories acquired from our tracking algorithm, we extract motion patterns from the trajectories
to detect certain activities performed by each person in the scene.
The motion patterns are defined in a simple rule-based manner.
Even though more complex methods such as variants of Hidden Markov Models \cite{oliver2000bayesian} to detect interactions 
could also be used,
our goal here is to demonstrate the usefulness of our tracking result and not test state-of-the-art interaction detection methods,
thus only a simple method was used.
The activities we detect are as follows:
\begin{itemize}
\item Room change: Given the tracking output, we can detect when someone enters or leaves a room. 
\item Sit down / stand up:
We trained a sitting detector \cite{voc-release5} which detects whether someone is sitting. 
Our algorithm looks for tracks which end/begin near a seat and check whether someone sat down/stood up 
around the same time.
\item Static interaction: If two people stand closer than distance $D'$ for duration $T'$, 
then it is likely that they are interacting. 
\item Dynamic interaction: If two people are moving with distance less than $D'$ apart for a duration longer than $T'$, 
and if they are moving faster than 20 cm/s, then it is highly likely that they are walking together. 
\end{itemize}
According to \cite{mcphail1982using},
if people are travelling in a group, then they should be at most 7 feet apart.
Therefore, we set the maximum distance $D'$ for there to be interaction between two people at 7 feet.
The minimum duration of interaction $T'$ was set to 8 seconds in our experiments.

Given the time and location of all the detected activities, 
we can sort the activities according to time and generate the visual diary.
The visual diary for a given individual consists of the following: 
\begin{itemize}
\item Snippets: snapshots and textual descriptions of the activity. 
Snapshots are extracted from video frames during the interaction and textual
descriptions are generated using natural language templates. 
\item Room/state timing estimates: time spent sitting or standing/walking in each room. 
\item Total interaction time: time spent in social interactions. 
\end{itemize}

Our proposed method of using tracking output for activity detection
can be easily combined with traditional activity recognition techniques using low-level features such as
Improved Dense Trajectories \cite{improvedtraj} with Fisher Vectors \cite{vggfisher}
to achieve better activity detection performance and detect more complex actions,
but extending activity recognition to activity detection
is beyond the scope of this paper.

\subsubsection*{Visual Diary Generation Results}
We performed long-term surveillance video summarization experiments by generating visual diaries on the
\emph{Caremedia 8h} sequence.
To acquire ground truth, we manually labeled the activities of three residents throughout the 
sequence. 
The nursing home residents were selected because they are the people we would like to focus on for the automatic analysis
of health status.
184 ground-truth activities were annotated.

We evaluated the different aspects of the visual diary: ``room/state timing estimates'', ``interaction timing estimates''
and ``snippet generation''.
The evaluation of ``room/state timing estimates'', i.e. predicted room location and state (sitting or upright), of a person
was done on the video frame level.
A frame was counted as true positive if the predicted state 
for a given video frame agrees with the ground truth.
False positives and false negatives were computed similarly.
To evaluate ``interaction timing estimates'', i.e. how much time a person spent in interactions,
a frame was only counted as true positive if 1) both the prediction result and ground truth result agree that
there was interaction and 2) the ID of the interacting targets match.
False positives and false negatives were computed similarly.
The evaluation of ``snippet generation'' accuracy was done as follows.
For snippets related to sit down, stand up and room change activities, a snippet was correct
if the predicted result and ground truth result had less than a 5 second time difference.
For social interaction-related snippets, a snippet was correct if more than 50\% of the predicted snippet contained a
matching ground truth interaction. 
Also, if a ground truth interaction was predicted as three separate interactions,
then only one interaction was counted as true positive while the other two were counted as false positives.
This prevents double counting of a single ground-truth interaction.

\begin{table}[tp]
\centering
	\begin{tabular}{ | @{\hspace{1pt}}C{3.3cm}@{\hspace{1pt}} || @{\hspace{1pt}}C{1.4cm}@{\hspace{1pt}} | @{\hspace{1pt}}C{1.4cm}@{\hspace{1pt}} | @{\hspace{1pt}}C{1.4cm}@{\hspace{1pt}} | }
		\hline
		Visual diary components	& Micro-Precision & Micro-Recall & Micro-F1	\\ \hline
		Snippet generation	& 0.382	& 0.522	& 0.441 \\ \hline
		Room/state timing estimates	& 0.809 & 0.511 & 0.626 \\ \hline
		Interaction timing estimates & 0.285 & 0.341  & 0.311 \\ \hline
	\end{tabular} 
\caption{Evaluation of generated visual diary.}
\label{tab:diary}
\end{table}

\begin{figure*}[tp]
\centering
\begin{subfigure}{0.48\textwidth}
\centering
\includegraphics[scale=0.28]{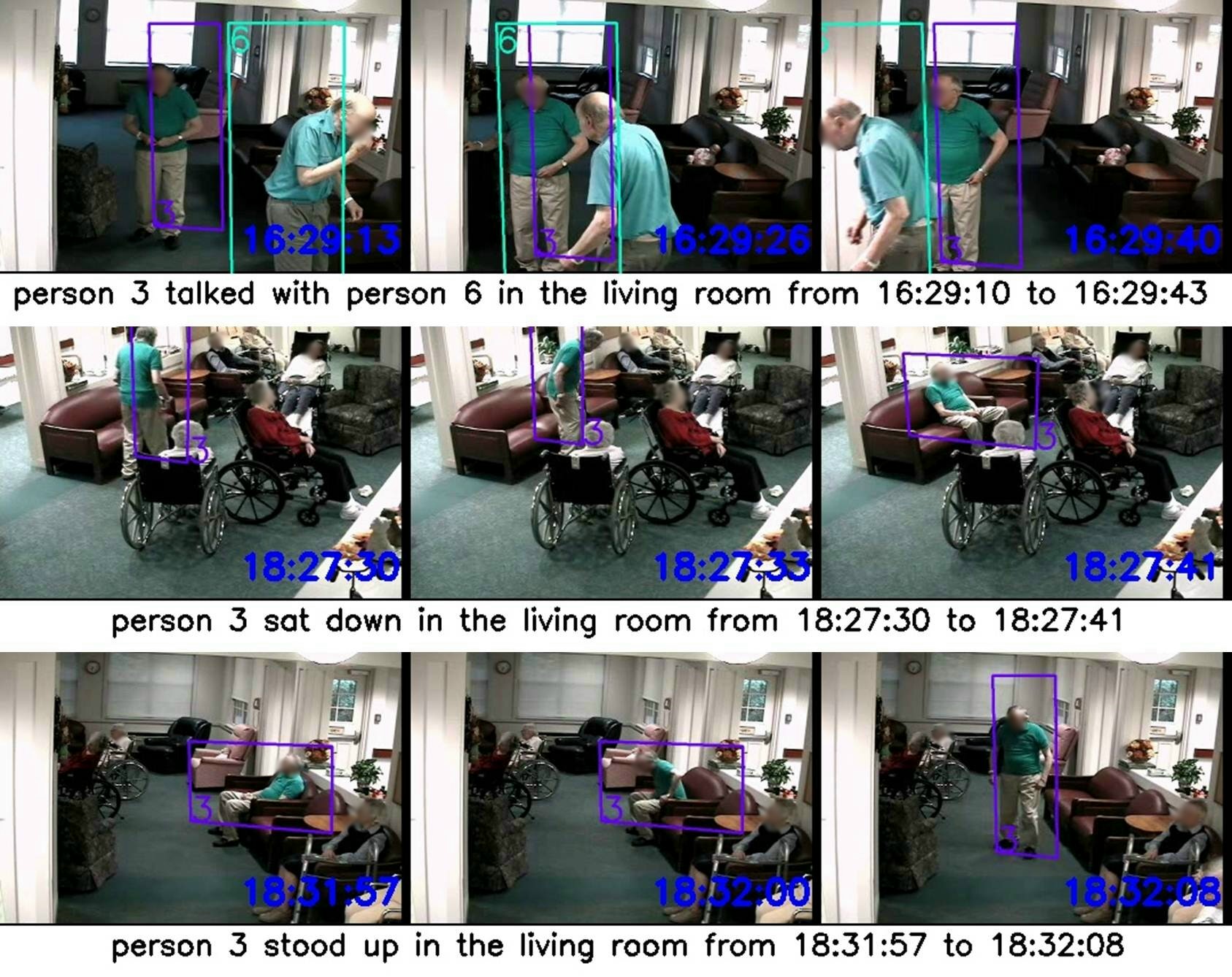}
\caption{Example snippets for resident 3.}\label{fig:snap3}
\end{subfigure}
\begin{subfigure}{0.48\textwidth}
\centering
\includegraphics[scale=0.28]{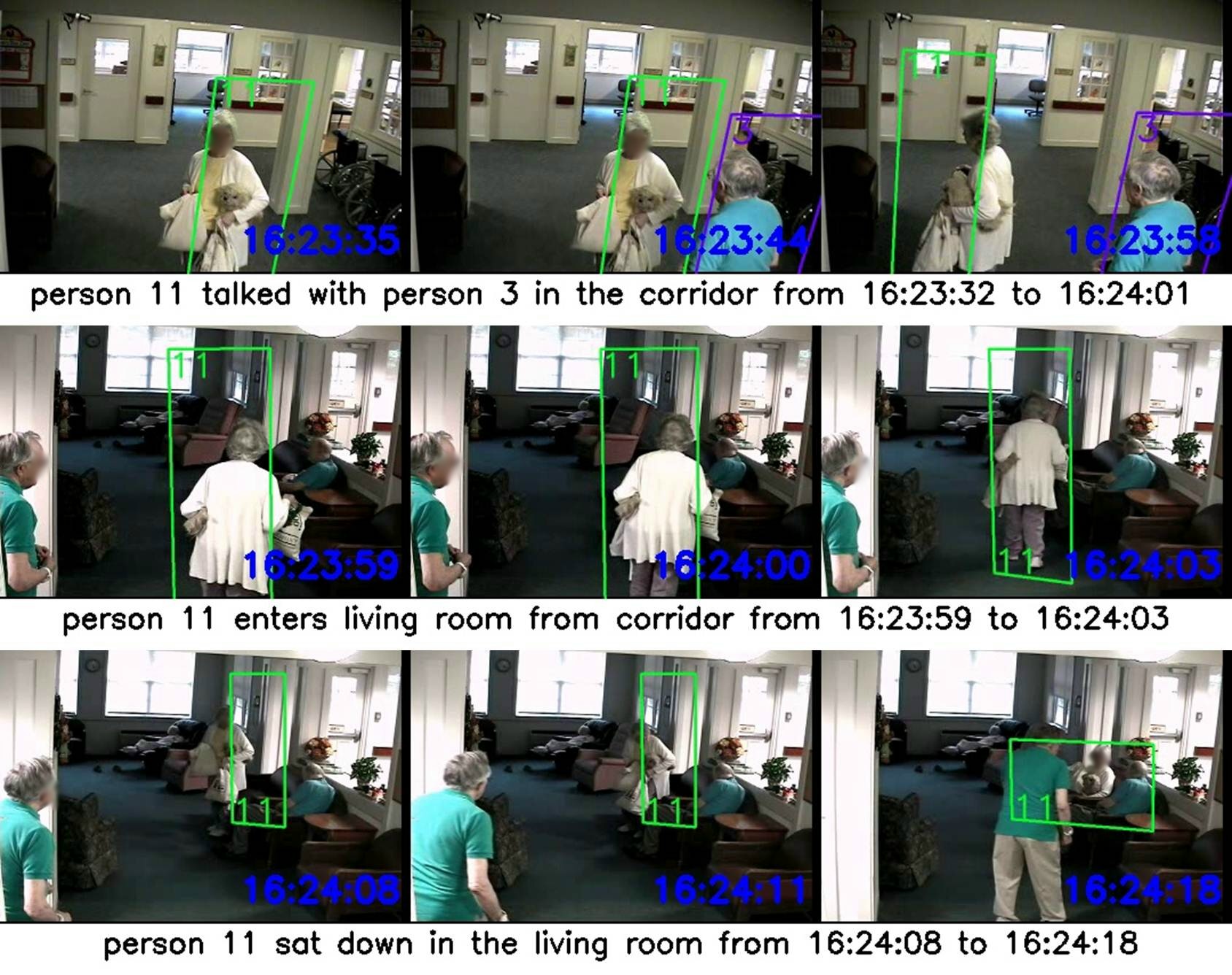}
\caption{Example snippets for resident 11.}\label{fig:snap6}
\end{subfigure}
\caption{
Example visual diary snippets for each resident.
}
\label{fig:diary}
\end{figure*}

Results are shown in Table~\ref{tab:diary}, which shows that
38\% of the generated snippets were correct,
and we have retrieved 52\% of the activities-of-interest.
For ``room/state timing estimates'', a 51.1\% recall shows that we know the state and room location of a person more than 50\% of the time.
The lower performance for ``interaction timing estimates'' was mainly caused by tracking failures, as both persons need to be
tracked correctly for interactions to be correctly detected and timings to be accurate.
These numbers are not high, but given that our method is fully automatic other than the collection of the face gallery,
this is a first cut at generating visual diaries for the elderly by
summarizing hundreds or even thousands of hours of surveillance video.

\begin{figure}[tp]
\centering
\includegraphics[scale=0.3]{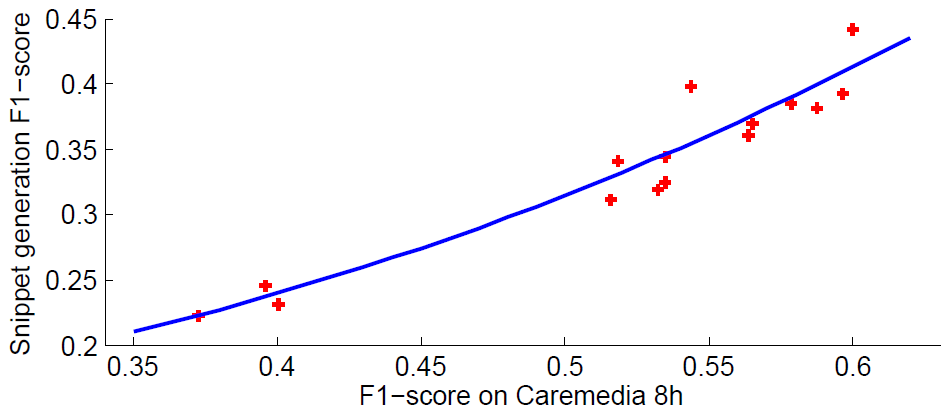}
\caption{
	Performance of snippet generation (y-axis) under varying tracking performance (x-axis).
}
\label{fig:diary_trend}
\end{figure}

We analyzed the effect of tracking performance
on snippet generation accuracy. 
We computed snippet generation F1-score for multiple tracking runs with varying tracking performance.
These runs include our baseline runs and also runs where we randomly corrupted face recognition labels to decrease tracking performance.
Results in Figure~\ref{fig:diary_trend} show that as tracking F1 increases,
snippet generation F1 also increases with a trend which could be fitted by a second-order polynomial.

Figure~\ref{fig:diary} shows example visual diaries for residents ID 3 and 11.
We can clearly see what each resident was doing at each time of the day.
Long term statistics shown in Figure~\ref{fig:diary11}
also clearly indicate the amount of time spent in each room and in social interactions.
If these statistics were computed over many days, 
a doctor or staff member could
start looking for patterns to better assess the status of health of a resident.

\section{Conclusion}
\label{sec:conclusions}

We present an identity-aware tracker which leverages face recognition information to enable automatic reinitialization of tracking targets 
in very long-term tracking scenarios.
Face recognition information is ideal in that it is robust to appearance and apparel change.
However, face recognition is unavailable in many frames, thus
we propagate identity information through a manifold learning framework which is solved by nonnegative matrix optimization.
Tracking experiments performed on up to 4,935 hours of video in a complex indoor environment showed that
our tracker was able to localize a person 53.2\% of the time with 69.8\% precision.
Accurate face recognition is key to good tracking results, where a 20\% increase in face recognition accuracy
will lead to around 10\% increase in tracking F1-score.
In addition to tracking experiments, we further utilized tracking output to 
generate visual diaries for identity-aware video summarization.
Experiments performed on 116.25 hours of video showed that
we can generate visual diary snippets with 38\% precision and 52\% recall.
Compared to tedious manual analysis of thousands of hours of surveillance video, 
our method is a strong alternative 
as it potentially opens the door to summarization of the ocean of surveillance video generated every day.

\ifCLASSOPTIONcaptionsoff
  \newpage
\fi



%
\bibliographystyle{IEEEtran} 
\bibliography{IEEEabrv,egbib} 
\begin{biography}[{\includegraphics[width=1in,height=1.25in,clip,keepaspectratio]{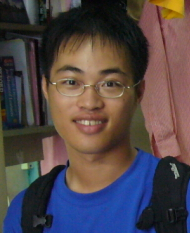}}]{Shoou-I Yu}
Shoou-I Yu received the B.S. in Computer Science and Information Engineering from National
Taiwan University, Taiwan in 2009. He is now a Ph.D. student in Language Technologies
Institute, Carnegie Mellon University. His research interests include multi-object tracking and multimedia retrieval.
\end{biography}
\begin{biography}[{\includegraphics[width=1in,height=1.25in,clip,keepaspectratio]{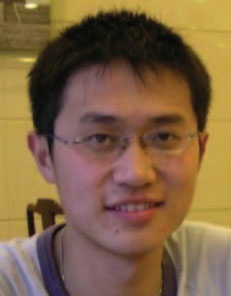}}]{Yi Yang}
Yi Yang received the PhD degree from Zhejiang University in 2010. He was a postdoc research fellow with the School of Computer Science at Carnegie Mellon University. He is now an Associate Professor with University of Technology Sydney. His research interest include multimedia, computer vision and machine learning.	
\end{biography}
\begin{biography}[{\includegraphics[width=1in,height=1.25in,clip,keepaspectratio]{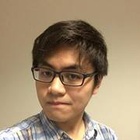}}]{Xuanchong Li}
Xuanchong Li received B.E. in computer science and technology from Zhejiang University, China in 2012. He is now a master student in Carnegie Mellon University. His research interest includes computer vision, machine learning.	
\end{biography}
\begin{biography}[{\includegraphics[width=1in,height=1.25in,clip,keepaspectratio]{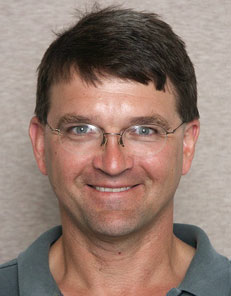}}]{Alexander G. Hauptmann}
Alexander G. Hauptmann received the B.A. and M.A. degrees in psychology from The Johns Hopkins University, Baltimore,
MD, USA, in 1982, the ``Diplom'' in computer science from the Technische Universit\"{a}t Berlin,Berlin, Germany, in 1984,
and the Ph.D. degree in computer science from Carnegie Mellon University (CMU), Pittsburgh, PA, USA in 1991. He is a Principal
Systems'' Scientist in the CMU Computer Science Department and also a faculty member 
with CMU's Language Technologies Institute. His research combines
the areas of multimedia analysis and retrieval, man-machine interfaces,
language processing, and machine learning. He is currently leading the
Informedia project which engages in understanding of video data ranging
from news to surveillance, Internet video for applications in general
retrieval as well as healthcare.

\end{biography}





\end{document}